\documentclass[10pt,twocolumn,letterpaper]{article}

\usepackage{iccv}
\usepackage{times}
\usepackage{epsfig}
\usepackage{graphicx}
\usepackage{amsmath}
\usepackage{amssymb}
\usepackage{url}
\usepackage{color}
\usepackage{amssymb}
\usepackage{bbm}
\usepackage{multirow}
\usepackage{etoolbox}
\usepackage{subcaption}

\newif\ifsupple
\newif\ifmain
\maintrue
\suppletrue

\usepackage[breaklinks=true,bookmarks=false]{hyperref}

\newcommand{\datasetname}{ActivityNet Captions}

\newcommand{\desc}[1]{\textit{#1}} 
\newcommand{\videoHours}{849 }
\newcommand{\totalVideos}{20k }
\newcommand{\totalSentences}{100k }

\newcommand{\averageSentencesPerVideo}{3.65 }
\newcommand{\stdDevSentencesPerVideo}{1.79 }
\newcommand{\averageWordsPerSentence}{13.48 }
\newcommand{\stdDevWordsPerSentence}{6.33 }
\newcommand{\averageWordsPerParagraph}{40 }
\newcommand{\stdDevWordsPerParagraph}{26 }

\newcommand{\datasetUrl}{\url{http://cs.stanford.edu/people/ranjaykrishna/densevid/}}

\iccvfinalcopy 

\def\iccvPaperID{271} 

\begin{document}

\title{Dense-Captioning Events in Videos}

\author{Ranjay Krishna, Kenji Hata, Frederic Ren, Li Fei-Fei, Juan Carlos Niebles\\
Stanford University\\
{\tt\small \{ranjaykrishna, kenjihata, fren, feifeili, jniebles\}@cs.stanford.edu}
}

\maketitle

\ifmain
\begin{abstract}
Most natural videos contain numerous events. For example, in a video of a ``man playing a piano'', the video might also contain ``another man dancing'' or ``a crowd clapping''. We introduce the task of dense-captioning events, which involves both detecting and describing events in a video. We propose a new model that is able to identify all events in a single pass of the video while simultaneously describing the detected events with natural language. Our model introduces a variant of an existing proposal module that is designed to capture both short as well as long events that span minutes. To capture the dependencies between the events in a video, our model introduces a new captioning module that uses contextual information from past and future events to jointly describe all events. We also introduce \datasetname, a large-scale benchmark for dense-captioning events. \datasetname~ contains \totalVideos videos amounting to \videoHours video hours with \totalSentences total descriptions, each with it's unique start and end time. Finally, we report performances of our model for dense-captioning events, video retrieval and localization.
\end{abstract}

\section{Introduction}
With the introduction of large scale activity datasets~\cite{laptev2008learning,karpathy2014large,THUMOS15,caba2015activitynet}, it has become possible to categorize videos into a discrete set of action categories~\cite{oneata2014efficient,gkioxari2015finding,gaidon2013temporal,wang2014video,tian2013spatiotemporal}. For example, in Figure~\ref{fig:pull_figure}, such models would output labels like \desc{playing piano} or \desc{dancing}. While the success of these methods is encouraging, they all share one key limitation: detail. To elevate the lack of detail from existing action detection models, subsequent work has explored explaining video semantics using sentence descriptions~\cite{pan2016jointly,rohrbach2015long,otani2016learning,venugopalan2014translating,venugopalan2015sequence}. For example, in Figure~\ref{fig:pull_figure}, such models would likely concentrate on \desc{an elderly man playing the piano in front of a crowd}. While this caption provides us more details about who is playing the piano and mentions an audience, it fails to recognize and articulate all the other events in the video. For example, at some point in the video, \desc{a woman starts singing along with the pianist} and then later \desc{another man starts dancing to the music}.  In order to identify all the events in a video and describe them in natural language, we introduce the task of \textit{dense-captioning events}, which requires a model to generate a set of descriptions for multiple events occurring in the video and localize them in time.

\begin{figure}[t]
  \centering
  \includegraphics[width=\columnwidth]{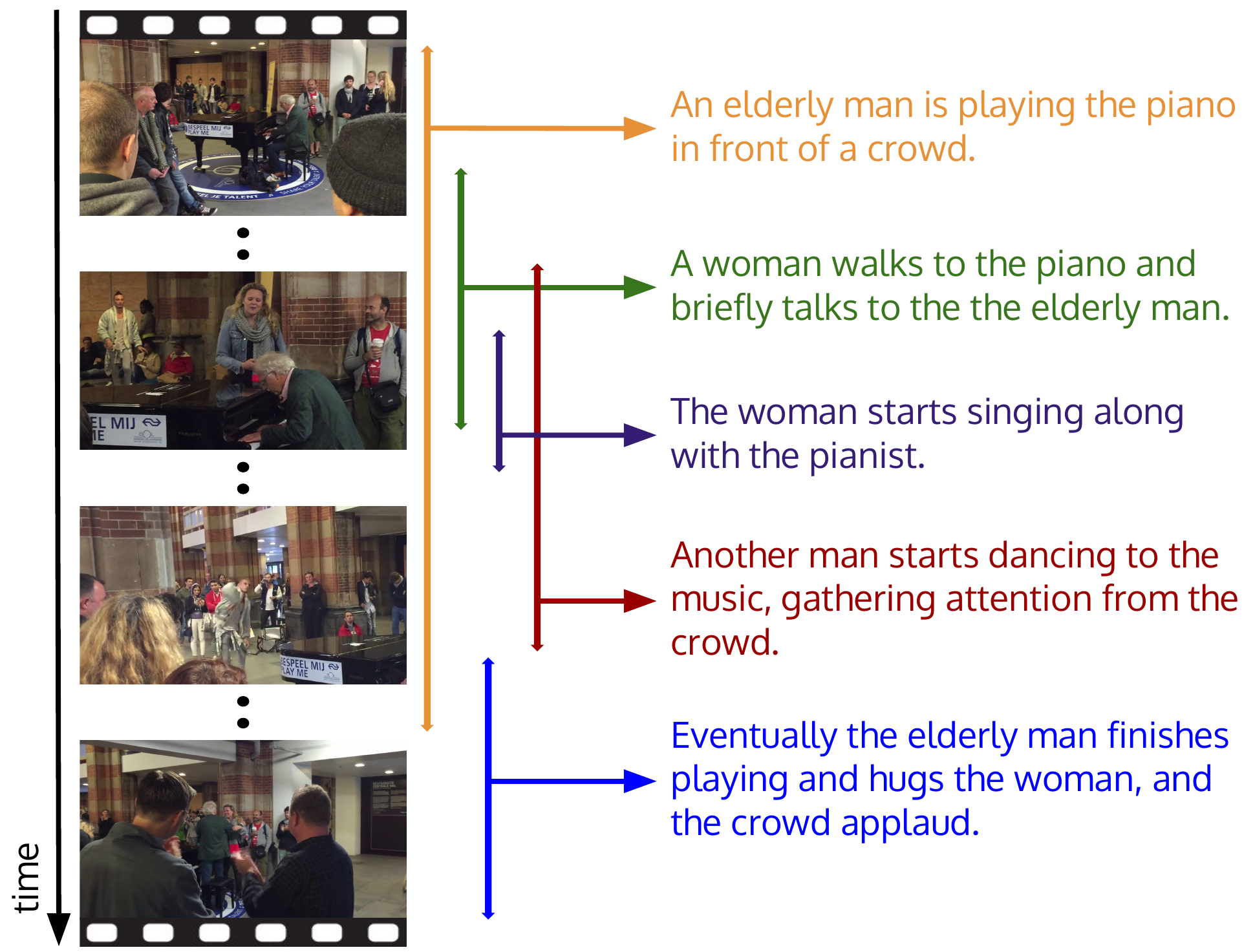}
\caption{Dense-captioning events in a video involves detecting multiple events that occur in a video and describing each event using natural language. These events are temporally localized in the video with independent start and end times, resulting in some events that might also occur concurrently and overlap in time.}
\label{fig:pull_figure}
\end{figure}

Dense-captioning events is analogous to dense-image-captioning~\cite{johnson2016densecap}; it describes videos and localize events in time whereas dense-image-captioning describes and localizes regions in space. However, we observe that dense-captioning events comes with its own set of challenges distinct from the image case. One observation is that events in videos can range across multiple time scales and can even overlap. While \desc{piano recitals} might last for the entire duration of a long video, \desc{the applause} takes place in a couple of seconds. To capture all such events, we need to design ways of encoding short as well as long sequences of video frames to propose events. Past captioning works have circumvented this problem by encoding the entire video sequence by mean-pooling~\cite{venugopalan2014translating} or by using a recurrent neural network (RNN)~\cite{venugopalan2015sequence}. While this works well for short clips, encoding long video sequences that span minutes leads to vanishing gradients, preventing successful training. To overcome this limitation, we extend recent work on generating action proposals~\cite{escorcia2016daps} to \textbf{multi-scale detection of events}. Also, our proposal module  processes each video in a forward pass, allowing us to detect events as they occur.

Another key observation is that the events in a given video are usually related to one another. In Figure~\ref{fig:pull_figure}, \desc{the crowd applauds} because a \desc{a man was playing the piano}. Therefore, our model must be able to use context from surrounding events to caption each event. A recent paper has attempted to describe videos with multiple sentences~\cite{yu2016video}. However, their model generates sentences for instructional ``cooking'' videos where the events occur sequentially and highly correlated to the objects in the video~\cite{rohrbach2014coherent}. We show that their model does not generalize to ``open'' domain videos where events are action oriented and can even overlap.
We introduce a \textbf{captioning module that utilizes the context} from all the events from our proposal module to generate each sentence. In addition, we show a variant of our captioning module that can operate on streaming videos by attending over only the past events. Our full model attends over both past as well as future events and demonstrates the importance of using context. 

To evaluate our model and benchmark progress in dense-captioning events, we introduce the \datasetname~ dataset\footnote{The dataset is available at \datasetUrl. For a detailed analysis of our dataset, please see our supplementary material.}. \datasetname~ contains \totalVideos videos taken from ActivityNet~\cite{caba2015activitynet}, where each video is annotated with a series of temporally localized descriptions (Figure~\ref{fig:pull_figure}). To showcase long term event detection, our dataset contains videos as long as $10$ minutes, with each video annotated with on average \averageSentencesPerVideo sentences. The descriptions refer to events that might be simultaneously occurring, causing the video segments to overlap. We ensure that each description in a given video is unique and refers to only one segment. While our videos are centered around human activities, the descriptions may also refer to non-human events such as: \desc{two hours later, the mixture becomes a delicious cake to eat}. We collect our descriptions using crowdsourcing find that there is high agreement in the temporal event segments, which is in line with research suggesting that brain activity is naturally structured into semantically meaningful events~\cite{baldassano2016discovering}.

With \datasetname, we are able to provide the first results for the task of dense-captioning events. Together with our online proposal module and our online captioning module, we show that we can detect and describe events in long or even streaming videos. We demonstrate that we are able to detect events found in short clips as well as in long video sequences. Furthermore, we show that utilizing context from other events in the video improves dense-captioning events. Finally, we demonstrate how \datasetname~ can be used to study video retrieval as well as event localization.

\section{Related work}
Dense-captioning events bridges two separate bodies of work: temporal action proposals and video captioning.
First, we review related work on action recognition, action detection and temporal proposals. Next, we survey how video captioning started from video retrieval and video summarization, leading to single-sentence captioning work. Finally, we contrast our work with recent work in captioning images and videos with multiple sentences.

Early work in \textbf{activity recognition} involved using hidden Markov models to learn latent action states~\cite{yamato1992recognizing}, followed by discriminative SVM models that used key poses and action grammars~\cite{niebles2010modeling,vahdat2011discriminative,pirsiavash2014parsing}. Similar works have used hand-crafted features~\cite{rohrbach2012database} or object-centric features~\cite{ni2014multiple} to recognize actions in fixed camera settings. More recent works have used dense trajectories~\cite{wang2014action} or deep learning features~\cite{karaman2014fast} to study actions. While our work is similar to these methods, we focus on describing such events with natural language instead of a fixed label set.

To enable action localization, \textbf{temporal action proposal} methods started from traditional sliding window approaches~\cite{duchenne2009automatic} and later started building models to propose a handful of possible action segments~\cite{escorcia2016daps,caba2016fast}. These proposal methods have used dictionary learning~\cite{caba2016fast} or RNN architectures~\cite{escorcia2016daps} to find possible segments of interest. However, such methods required each video frame to be processed once for every sliding window. DAPs introduced a framework to allow proposing overlapping segments using a sliding window. We modify this framework by removing the sliding windows and outputting proposals at every time step in a single pass of the video. We further extend this model and enable it to detect long events by implementing a multi-scale version of DAPs, where we sample frames at longer strides.

Orthogonal to work studying proposals, early approaches that connected video with language studied the task of \textbf{video retrieval with natural language}. They worked on generating a common embedding space between language and videos~\cite{otani2016learning,xu2015jointly}. Similar to these, we evaluate how well existing models perform on our dataset. Additionally, we introduce the task of localizing a given sentence given a video frame, allowing us to now also evaluate whether our models are able to locate specified events.

In an effort to start describing videos, methods in \textbf{video summarization} aimed to congregate segments of videos that include important or interesting visual information~\cite{yao2016highlight,yang2015unsupervised,gygli2015video,boiman2007detecting}. These methods attempted to use low level features such as color and motion or attempted to model objects~\cite{zhang1997integrated} and their relationships~\cite{wolf1996key,goldman2006schematic} to select key segments. Meanwhile, others have utilized text inputs from user studies to guide the selection process~\cite{song2015tvsum,liu2015multi}. While these summaries provide a means of finding important segments, these methods are limited by small vocabularies and do not evaluate how well we can explain visual events~\cite{yeung2014videoset}. 

After these summarization works, early attempts at \textbf{video captioning}~\cite{venugopalan2014translating} simply mean-pooled video frame features and used a pipeline inspired by the success of image captioning~\cite{karpathy2015deep}. However, this approach only works for short video clips with only one major event. To avoid this issue, others have proposed either a recurrent encoder~\cite{donahue2015long,venugopalan2015sequence,xu2015multi} or an attention mechanism~\cite{yao2015describing}. To capture more detail in videos, a new paper has recommended describing videos with paragraphs (a list of sentences) using a hierarchical RNN~\cite{mikolov2010recurrent} where the top level network generates a series of hidden vectors that are used to initialize low level RNNs that generate each individual sentence~\cite{yu2016video}. While our paper is most similar to this work, we address two important missing factors. First, the sentences that their model generates refer to different events in the video but are not localized in time. Second, they use the TACoS-MultiLevel~\cite{rohrbach2014coherent}, which contains less than $200$ videos and is constrained to ``cooking'' videos and only contain non-overlapping sequential events. We address these issues by introducing the \datasetname~ dataset which contains overlapping events and by introducing our captioning module that uses temporal context to capture the interdependency between all the events in a video.

Finally, we build upon the recent work on \textbf{dense-image-captioning}~\cite{johnson2016densecap}, which generates a set of localized descriptions for an image. Further work for this task has used spatial context to improve captioning~\cite{yang2016dense,xu2015show}. Inspired by this work, and by recent literature on using spatial attention to improve human tracking~\cite{alahi2016social}, we design our captioning module to incorporate temporal context (analogous to spatial context except in time) by attending over the other events in the video.

\section{Dense-captioning events model}
\label{sec:model}

\begin{figure*}[t]
  \centering
  \includegraphics[width=0.95\textwidth]{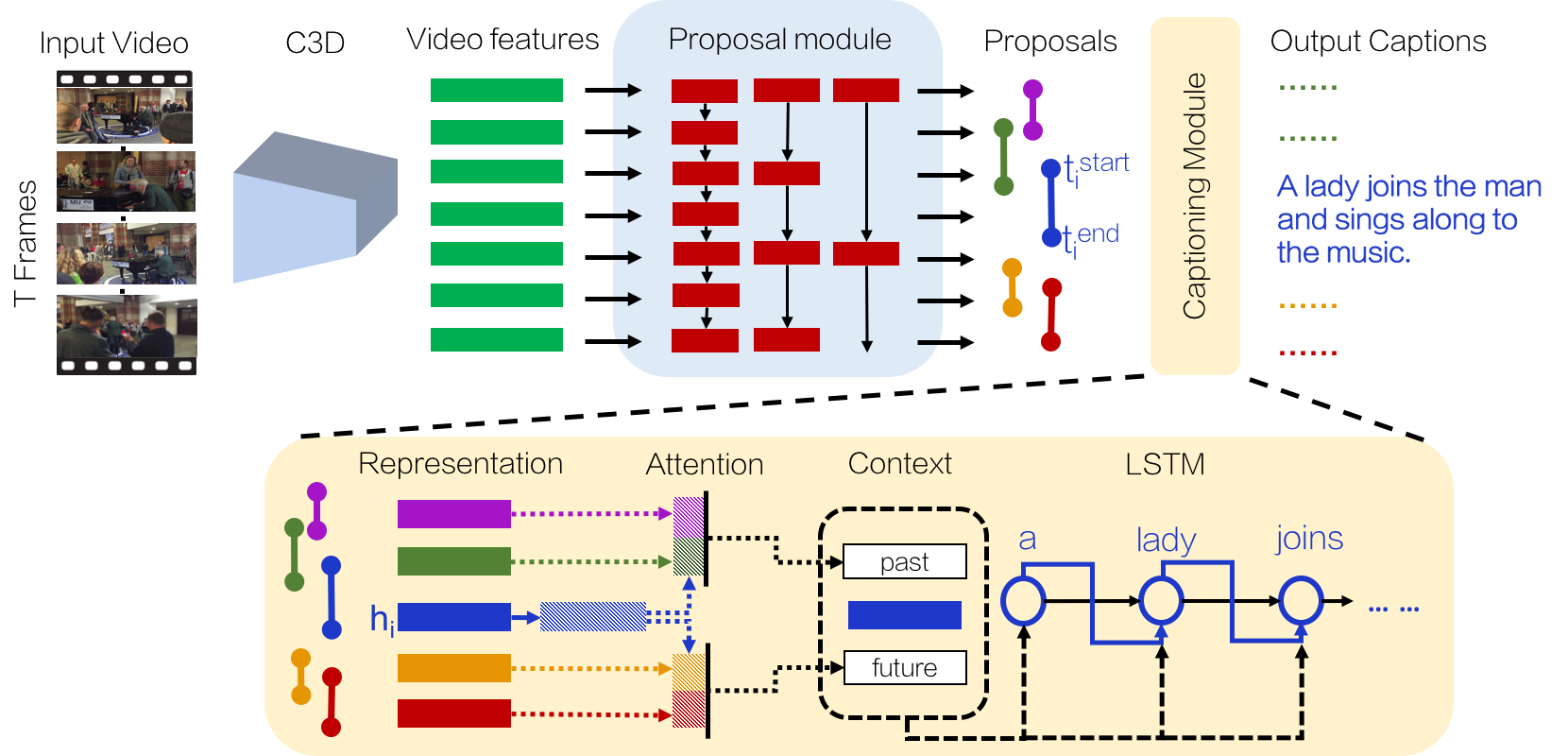}
\caption{Complete pipeline for dense-captioning events in videos with  descriptions. We first extract C3D features from the input video. These features are fed into our proposal module at varying stride to predict both short as well as long events. Each proposal, which consists of a unique start and end time and a hidden representation, is then used as input into the captioning module. Finally, this captioning model leverages context from neighboring events to generate each event description.}
\label{fig:model}
\end{figure*}

\paragraph{Overview.} Our goal is to design an architecture that jointly localizes temporal proposals of interest and then describes each with natural language. The two main challenges we face are to develop a method that can (1) detect multiple events in short as well as long video sequences and (2) utilize the context from past, concurrent and future events to generate descriptions of each one. Our proposed architecture (Figure~\ref{fig:model}) draws on architectural elements present in recent work on action proposal~\cite{escorcia2016daps} and social human tracking~\cite{alahi2016social} to tackle both these challenges. 

Formally, the input to our system is a sequence of video frames $v = \{v_t\}$ where $t \in {0, ..., T-1}$ indexes the frames in temporal order. Our output is a set of sentences $s_i \in \mathcal{S}$ where $s_i = (t^{\mathrm{start}}, t^{\mathrm{end}}, \{v_j\})$ consists of the start and end times for each sentence which is defined by a set of words $v_j \in V$ with differing lengths for each sentence and $V$ is our vocabulary set. 

Our model first sends the video frames through a proposal module that generates a set of proposals:
\begin{equation}
P = \{(t_i^{\mathrm{start}}, t_i^{\mathrm{end}}, \mathrm{score}_i, h_i)\}
\end{equation}
All the proposals with a $score_i$ higher than a threshold are forwarded to our language model that uses context from the other proposals while captioning each event. The hidden representation $h_i$ of the event proposal module is used as inputs to the captioning module, which then outputs descriptions for each event, while utilizing the context from the other events.

\subsection{Event proposal module}
The proposal module in Figure~\ref{fig:model} tackles the challenge of detecting events in short as well as long video sequences, while preventing the dense application of our language model over sliding windows during inference. Prior work usually pools video features globally into a fixed sized vector~\cite{donahue2015long,venugopalan2015sequence,xu2015multi}, which is sufficient for representing short video clips but is unable to detect multiple events in long videos. Additionally, we would like to detect events in a single pass of the video so that the gains over a simple temporal sliding window are significant. To tackle this challenge, we design an event proposal module to be a variant of DAPs~\cite{escorcia2016daps} that can detect longer events. 

\noindent \textbf{Input.} Our proposal module receives a series of features capturing semantic information from the video frames. Concretely, the input to our proposal module is a sequence of features: 
$\{f_t = F(v_t:v_{t+\delta})\}$
where $\delta$ is the time resolution of each feature $f_t$. In our paper, $F$ extracts C3D features~\cite{ji20133d} where $\delta=16$ frames. The output of $F$ is a tensor of size $N{\times}D$ where $D=500$ dimensional features and $N=T/\delta$ discretizes the video frames.

\noindent \textbf{DAPs.} Next, we feed these features into a variant of DAPs~\cite{escorcia2016daps} where we sample the videos features at different strides ($1$, $2$, $4$ and $8$ for our experiments) and feed them into a proposal long short-term memory (LSTM) unit. The longer strides are able to capture longer events. The LSTM accumulates evidence across time as the video features progress. We do not modify the training of DAPs and only change the model at inference time by outputting $K$ proposals at every time step, each proposing an event with offsets. So, the LSTM is capable of generating proposals at different overlapping time intervals and we only need to iterate over the video once, since all the strides can be computed in parallel. Whenever the proposal LSTM detects an event, we use the hidden state of the LSTM at that time step as a feature representation of the visual event. Note that the proposal model can output proposals for events that can be overlapping. While traditional DAPs uses non-maximum suppression to eliminate overlapping outputs, we keep them separately and treat them as individual events.

\subsection{Captioning module with context}
Once we have the event proposals, the next stage of our pipeline is responsible for describing each event. A naive captioning approach could treat each description individually and use a captioning LSTM network to describe each one. However, most events in a video are correlated and can even cause one another. For example, we saw in Figure~\ref{fig:pull_figure} that the \desc{man playing the piano} caused the \desc{other person to start dancing}. We also saw that after the man finished playing the piano, the \desc{audience applauded}. 
To capture  such correlations, we design our captioning module to incorporate the ``context'' from its neighboring events. Inspired by recent work~\cite{alahi2016social} on human tracking that utilizes spatial context between neighboring tracks, we develop an analogous model that captures temporal context in videos by grouping together events in time instead of tracks in space.
	
\noindent \textbf{Incorporating context.}
To capture the context from all other neighboring events, we categorize all events into two buckets relative to a reference event. These two context buckets capture events that have already occurred (past), and events that take place after this event has finished (future). Concurrent events are split into one of the two buckets: past if it end early and future otherwise. For a given video event from the proposal module, with hidden representation $h_i$ and start and end times of $[t^{\mathrm{start}}_i, t^{\mathrm{end}}_i]$, we calculate the past and future context representations as follows:

\begin{align}
h^{\mathrm{past}}_i = \frac{1}{Z^{\mathrm{past}}} \sum_{j \ne i} \mathbbm{1}[t^{\mathrm{end}}_j < t^{\mathrm{end}}_i] w_j h_j \\
h^{\mathrm{future}}_i = \frac{1}{Z^{\mathrm{future}}} \sum_{j \ne i} \mathbbm{1}[t^{\mathrm{end}}_j >= t^{\mathrm{end}}_i] w_j h_j
\end{align}
where $h_j$ is the hidden representation of the other proposed events in the video. $w_j$ is the weight used to determine how relevant event $j$ is to event $i$. $Z$ is the normalization that is calculated as $Z^{\mathrm{past}} = \sum_{j \ne i} \mathbbm{1}[t^{\mathrm{end}}_j < t^{\mathrm{end}}_i]$. We calculate $w_j$ as follows:

\begin{align}
a_i = w_a h_i + b_a \\
w_j = a_i h_j
\label{eq:context}
\end{align}
where $a_i$ is the attention vector calculated from the learnt weights $w_a$ and bias $b_a$. We use the dot product of $a_i$ and $h_j$ to calculate $w_j$. The concatenation of $(h^{\mathrm{past}}_i, h_i, h^{\mathrm{future}}_i)$ is then fed as the input to the captioning LSTM that describes the event. With the help of the context, each LSTM also has knowledge about events that have happened or will happen and can tune its captions accordingly.

\noindent \textbf{Language modeling.} Each language LSTM is initialized to have $2$ layers with $512$ dimensional hidden representation. We randomly initialize all the word vector embeddings from a Gaussian with standard deviation of $0.01$. We sample predictions from the model using beam search of size $5$.

\subsection{Implementation details.}

\noindent \textbf{Loss function.} We use two separate losses to train both our proposal model ($\mathcal{L}_{\mathrm{prop}}$) and our captioning model ($\mathcal{L}_{\mathrm{cap}}$). Our proposal models predicts confidences ranging between $0$ and $1$ for varying proposal lengths. We use a weighted cross-entropy term to evaluate each proposal confidence. 

We only pass to the language model proposals that have a high IoU with ground truth proposals. Similar to previous work on language modeling~\cite{krause2016paragraphs,karpathy2015deep}, we use a cross-entropy loss across all words in every sentence. We normalize the loss by the batch-size and sequence length in the language model. We weight the contribution of the captioning loss with $\lambda_1=1.0$ and the proposal loss with $\lambda_2=0.1$:

\begin{equation}
\mathcal{L} = \lambda_1 \mathcal{L}_{\mathrm{cap}} + \lambda_2 \mathcal{L}_{\mathrm{prop}}
\end{equation}

\noindent \textbf{Training and optimization.} We train our full dense-captioning model by alternating between training the language model and the proposal module every $500$ iterations. We first train the captioning module by masking all neighboring events for $10$ epochs before adding in the context features. We initialize all weights using a Gaussian with standard deviation of $0.01$. We use stochastic gradient descent with momentum $0.9$ to train. We use an initial learning rate of $1{\times}10^{-2}$ for the language model and $1{\times} 10^{-3}$ for the proposal module. For efficiency, we do not finetune the C3D feature extraction.

Our training batch-size is set to $1$. We cap all sentences to be a maximum sentence length of $30$ words and implement all our code in PyTorch 0.1.10. One mini-batch runs in approximately $15.84$ ms on a Titan X GPU and it takes 2 days for the model to converge.

\begin{figure}[t]
  \centering
  \includegraphics[width=\columnwidth]{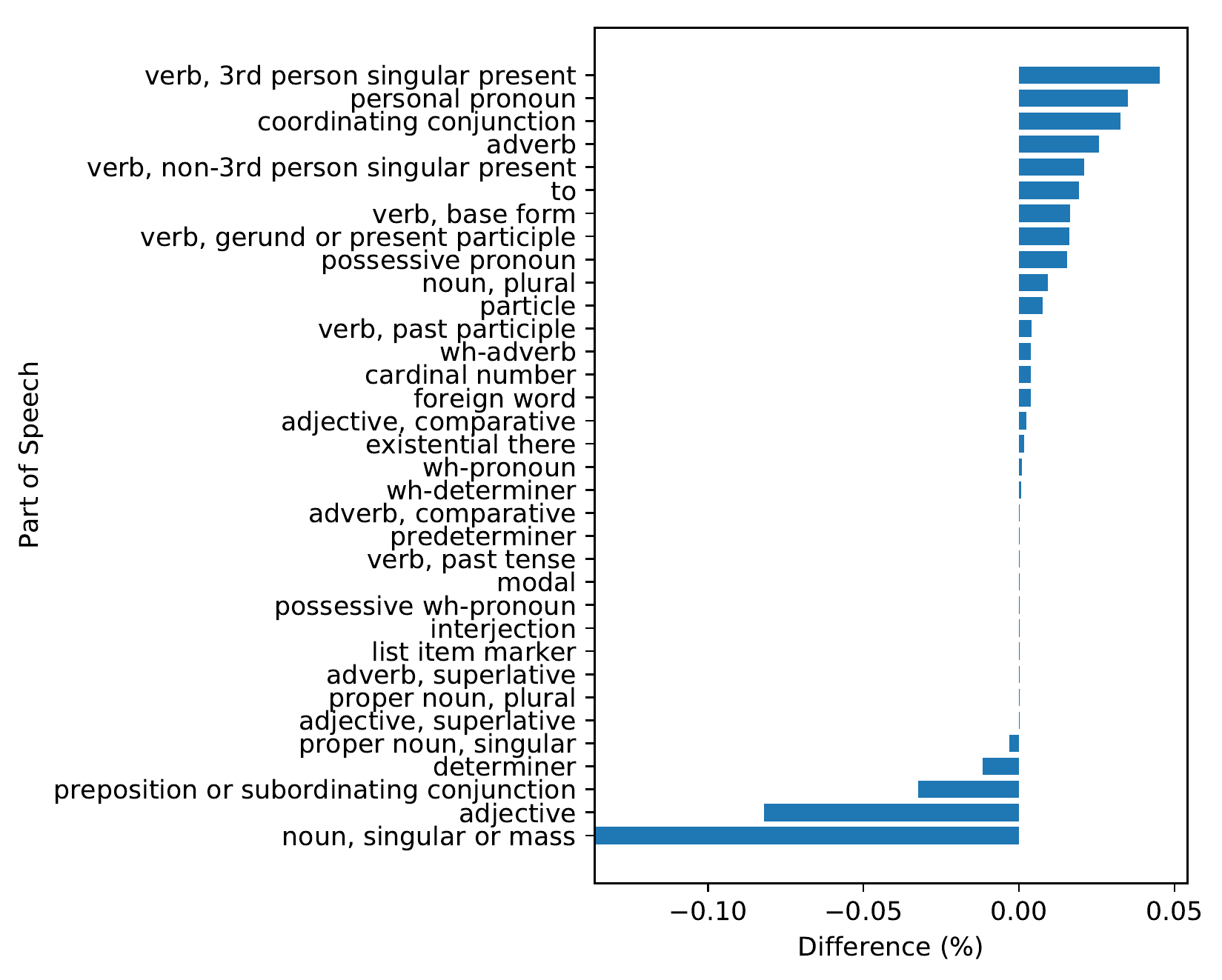}
\caption{The parts of speech distribution of \datasetname~ compared with Visual Genome, a dataset with multiple sentence annotations per image. There are many more verbs and pronouns represented in \datasetname, as the descriptions often focus on actions.}
\label{fig:part_of_speech}
\end{figure}

\section{\datasetname~ dataset}
\label{sec:dataset}
The \datasetname~ dataset connects videos to a series of temporally annotated sentences. Each sentence covers an unique segment of the video, describing an event that occurs. These events may occur over very long or short periods of time and are not limited in any capacity, allowing them to co-occur. We will now present an overview of the dataset and also provide a detailed analysis and comparison with other datasets in our supplementary material.

\begin{table*}[t]
\centering
\begin{tabular}{l c c c c c c | c c c c c c}
& \multicolumn{6}{c|}{\textbf{with GT proposals}} & \multicolumn{6}{c}{\textbf{with learnt proposals}} \\
& B@1 & B@2 & B@3 & B@4 & M & C & B@1 & B@2 & B@3 & B@4 & M & C \\ 
\hline \hline
LSTM-YT~\cite{venugopalan2015sequence} & 18.22 & 7.43 & 3.24 & 1.24 & 6.56 & 14.86 & - & - & - & - & - & - \\
S2VT~\cite{venugopalan2014translating} & 20.35 & 8.99 & 4.60 & 2.62 & 7.85 & 20.97 & - & - & - & - & - & - \\
H-RNN~\cite{yu2016video}               & 19.46 & 8.78 & 4.34 & 2.53 & 8.02 & 20.18 & - & - & - & - & - & - \\
\hline
no context (ours)  & 20.35 & 8.99  & 4.60 & 2.62 & 7.85 & 20.97 & 12.23 & 3.48 & 2.10 & 0.88 & 3.76 & 12.34 \\
online$-$attn (ours) & 21.92 & 9.88  & 5.21 & 3.06 & 8.50 & 22.19 & 15.20 & 5.43 & 2.52 & 1.34 & 4.18 & 14.20 \\
online (ours)      & 22.10 & 10.02 & 5.66 & 3.10 & 8.88 & 22.94 & 17.10 & 7.34 & 3.23 & 1.89 & 4.38 & 15.30 \\
full$-$attn (ours)   & 26.34 & 13.12 & 6.78 & 3.87 & 9.36 & 24.24 & 15.43 & 5.63 & 2.74 & 1.72 & 4.42 & 15.29 \\
full (ours) & \textbf{26.45} & \textbf{13.48} & \textbf{7.12} & \textbf{3.98} & \textbf{9.46} & \textbf{24.56} & \textbf{17.95} & \textbf{7.69} & \textbf{3.86} & \textbf{2.20} & \textbf{4.82} & \textbf{17.29} \\
\end{tabular}
\caption{We report Bleu (B), METEOR (M) and CIDEr (C) captioning scores for the task of dense-captioning events. On the left, we report performances of just our captioning module with ground truth proposals. On the right, we report the combined performances of our complete model, with proposals predicted from our proposal module. Since prior work has focused only on describing entire videos and not also detecting a series of events, we only compare existing video captioning models using ground truth proposals.}
\label{tab:multicaption}
\end{table*}

\subsection{Dataset statistics}
On average, each of the \totalVideos videos in \datasetname~ contains \averageSentencesPerVideo temporally localized sentences, resulting in a total of \totalSentences sentences. We find that the number of sentences per video follows a relatively normal distribution. Furthermore, as the video duration increases, the number of sentences also increases. Each sentence has an average length of \averageWordsPerSentence words, which is also normally distributed.

On average, each sentence describes $36$ seconds and $31\%$ of their respective videos. However, the entire paragraph for each video on average describes $94.6\%$ of the entire video, demonstrating that each paragraph annotation still covers all major actions within the video. Furthermore, we found that $10\%$ of the temporal descriptions overlap, showing that the events cover simultaneous events.

Finally, our analysis on the sentences themselves indicate that \datasetname~ focuses on verbs and actions. In Figure~\ref{fig:part_of_speech}, we compare against Visual Genome~\cite{krishnavisualgenome}, the image dataset with most number of image descriptions (\~4.5 million). With the percentage of verbs comprising \datasetname being significantly more, we find that \datasetname~ shifts sentence descriptions from being object-centric in images to action-centric in videos. Furthermore, as there exists a greater percentage of pronouns in \datasetname, we find that the sentence labels will more often refer to entities found in prior sentences.


\subsection{Temporal agreement amongst annotators}
To verify that \datasetname~'s captions mark semantically meaningful events~\cite{baldassano2016discovering}, we collected two distinct, temporally annotated paragraphs from different workers for each of the $4926$ validation and $5044$ test videos. Each pair of annotations was then tested to see how well they temporally corresponded to each other. We found that, on average, each sentence description had an tIoU of $70.2\%$ with the maximal overlapping combination of sentences from the other paragraph. Since these results agree with prior work~\cite{baldassano2016discovering}, we found that workers generally agree with each other when annotating temporal boundaries of video events.


\section{Experiments}
\label{sec:experiments}
We evaluate our model by detecting multiple events in videos and describing them. We refer to this task as dense-captioning events (Section~\ref{sec:multi-captioning}). We test our model on \datasetname, which was built specifically for this task.

Next, we provide baseline results on two additional tasks that are possible with our model. The first of these tasks is localization (Section~\ref{sec:localization}), which tests our proposal model's capability to adequately localize all the events for a given video. The second task is retrieval (Section~\ref{sec:retrieval}), which tests a variant of our model's ability to recover the correct set of sentences given the video or vice versa. Both these tasks are designed to test the event proposal module (localization) and the captioning module (retrieval) individually.

\subsection{Dense-captioning events}
\label{sec:multi-captioning}
To dense-caption events, our model is given an input video and is tasked with detecting individual events and describing each one with natural language.

\noindent \textbf{Evaluation metrics.} Inspired by the dense-image-captioning~\cite{johnson2016densecap} metric, we use a similar metric to measure the joint ability of our model to both localize and caption events. This metric computes the average precision across tIoU thresholds of $0.3$, $0.5$, $0.7$ when captioning the top $1000$ proposals. We measure precision of our captions using traditional evaluation metrics: Bleu, METEOR and CIDEr.

To isolate the performance of language in the predicted captions without localization, we also use ground truth locations across each test image and evaluate predicted captions.

\begin{table}[t]
\centering
\begin{tabular}{l c c c c c c}
& B@1 & B@2 & B@3 & B@4 & M & C \\ 
\hline \hline
\multicolumn{7}{l}{\textbf{no context}}\\
$1^{st}$ sen. & 23.60 & 12.19 & 7.11 & 4.51 & 9.34 & \textbf{31.56} \\
$2^{nd}$ sen. & 19.74 & 8.17 & 3.76 & 1.87 & 7.79 & 19.37 \\
$3^{rd}$ sen. & 18.89 & 7.51 & 3.43 & 1.87 & 7.31 & 19.36\\
\hline
\multicolumn{7}{l}{\textbf{online}}\\
$1^{st}$ sen. & 24.93 & 12.38 & 7.45 & 4.77 & 8.10 & 30.92 \\
$2^{nd}$ sen. & 19.96 & 8.66 & 4.01 & 1.93 & 7.88 & 19.17 \\
$3^{rd}$ sen. & 19.22 & 7.72 & 3.56 & \textbf{1.89} & 7.41 & 19.36 \\
\hline
\multicolumn{7}{l}{\textbf{full}}\\
$1^{st}$ sen. & \textbf{26.33} & \textbf{13.98} & \textbf{8.45} & \textbf{5.52} & \textbf{10.03} & 29.92 \\
$2^{nd}$ sen. & \textbf{21.46} & \textbf{9.06} & \textbf{4.40} & \textbf{2.33} & \textbf{8.28} & \textbf{20.17} \\
$3^{rd}$ sen. & \textbf{19.82} & \textbf{7.93} & \textbf{3.63} & 1.83 & \textbf{7.81} & \textbf{20.01} \\
\end{tabular}
\caption{We report the effects of context on captioning the $1^{st}$, $2^{nd}$ and $3^{rd}$ events in a video. We see that performance increases with the addition of past context in the online model and with future context in full model.}
\label{tab:sents}
\end{table}

\begin{figure}[t]
  \centering
  \begin{subfigure}[a]{\columnwidth}
        \centering
        \includegraphics[width=\columnwidth]{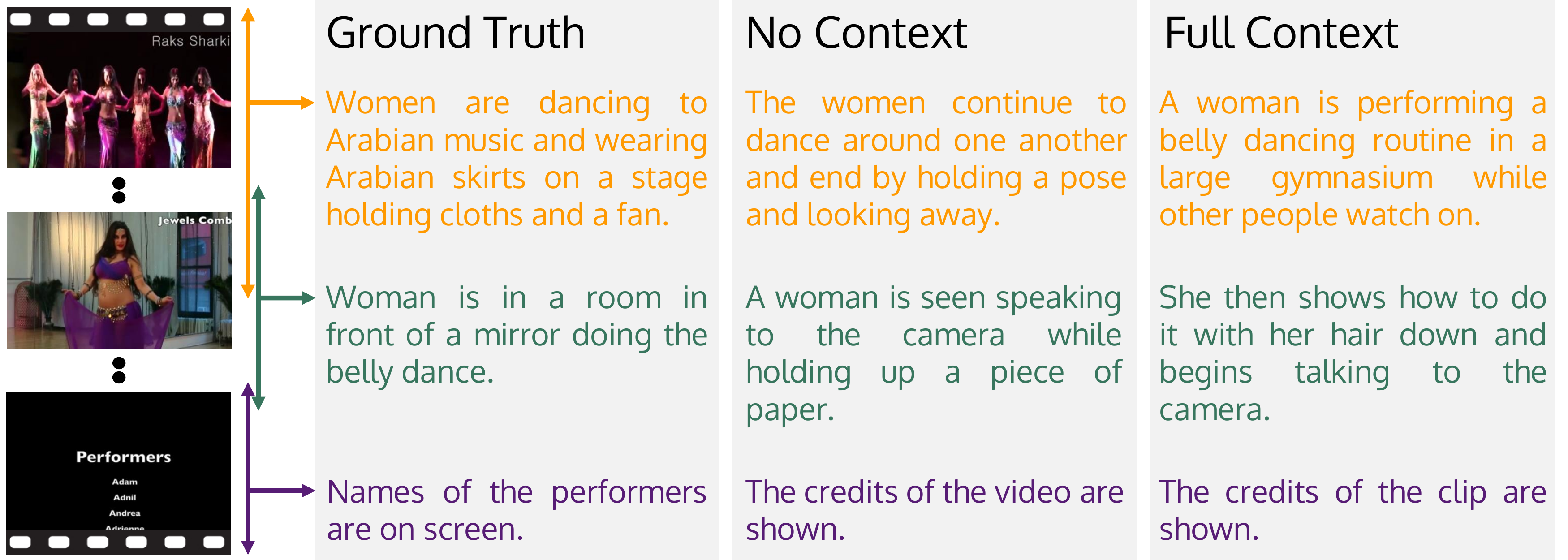}
        \caption{Adding context can generate consistent captions.}
   \end{subfigure}
   \begin{subfigure}[b]{\columnwidth}
        \centering
        \includegraphics[width=\columnwidth]{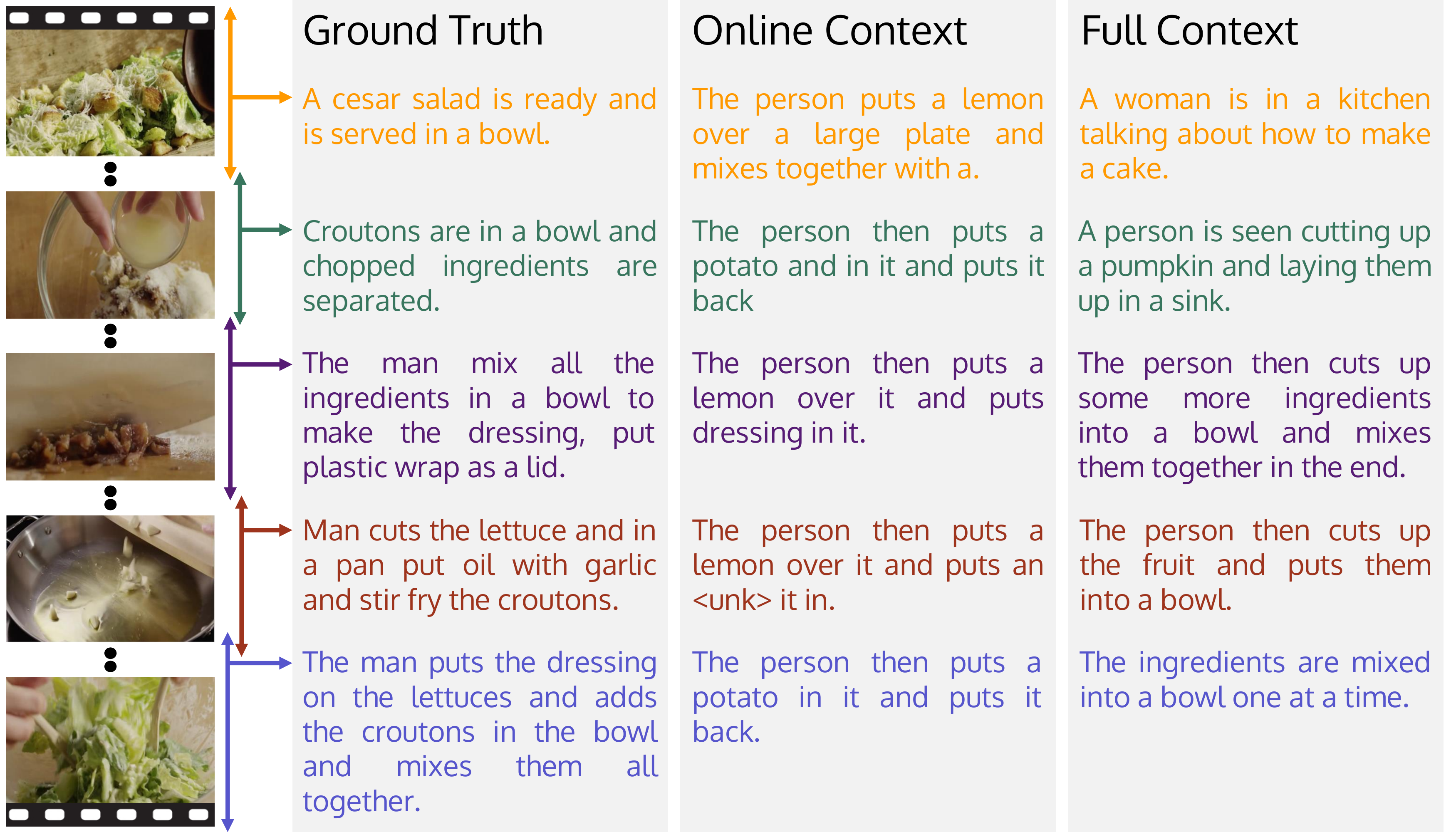}
        \caption{Comparing \textit{online} versus \textit{full} model.}
   \end{subfigure}
   \begin{subfigure}[c]{\columnwidth}
        \centering
        \includegraphics[width=\columnwidth]{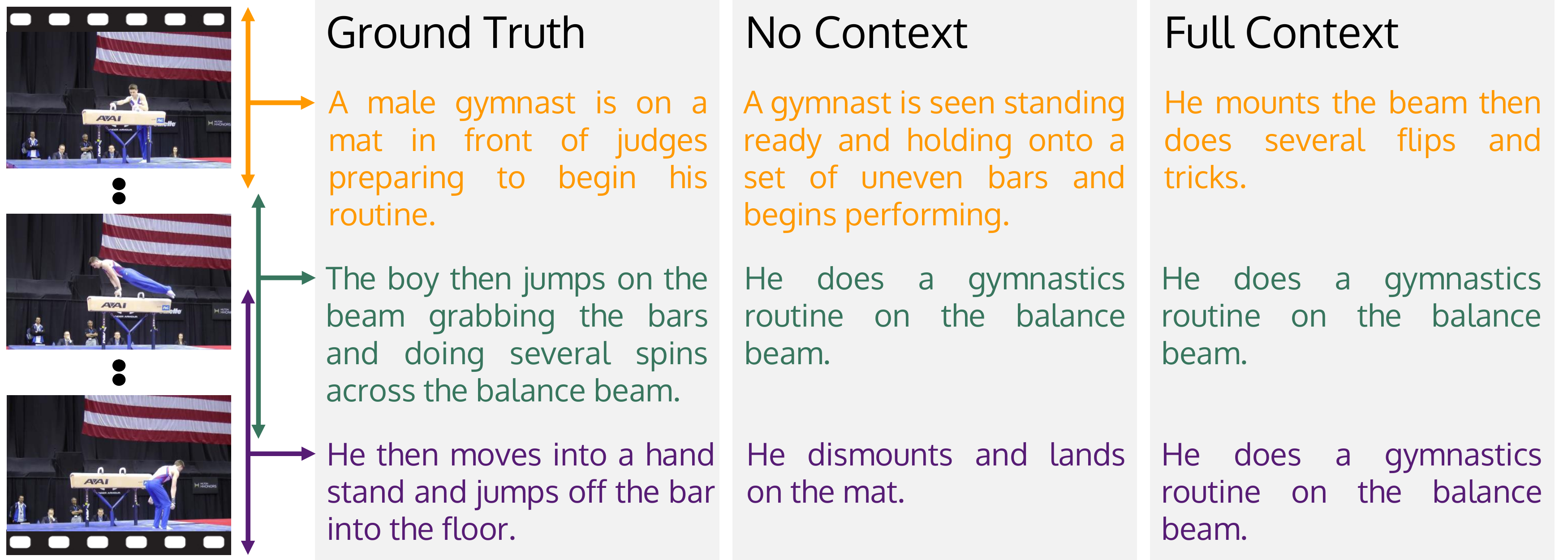}
        \caption{Context might add more noise to rare events.}
   \end{subfigure}
  
\caption{Qualitative dense-captioning captions generated using our model. We show captions with the highest overlap with ground truth captions.}
\label{fig:qualitative_results}
\end{figure}

\noindent \textbf{Baseline models.} Since all the previous models proposed so far have focused on the task of describing entire videos and not detecting a series of events, we only compare existing video captioning models using ground truth proposals. Specifically, we compare our work with \textit{LSTM-YT}~\cite{venugopalan2015sequence}, \textit{S2VT}~\cite{venugopalan2014translating} and \textit{H-RNN}~\cite{yu2016video}. \textit{LSTM-YT} pools together video features to describe videos while \textit{S2VT}~\cite{venugopalan2014translating} encodes a video using an RNN. \textit{H-RNN}~\cite{yu2016video} generates paragraphs by using one RNN to caption individual sentences while the second RNN is used to sequentially initialize the hidden state for the next sentence generation. Our model can be though of as a generalization of the \textit{H-RNN} model as it uses context, not just from the previous sentence but from surrounding events in the video. Additionally, our method treats context, not as features from object detectors but encodes it from unique parts of the proposal module.

\noindent \textbf{Variants of our model.} Additionally, we compare different variants of our model. Our \textit{no context} model is our implementation of \textit{S2VT}. The \textit{full} model is our complete model described in Section~\ref{sec:model}. The \textit{online} model is a version of our full model that uses context only from past events and not from future events. This version of our model can be used to caption long streams of video in a single pass. The \textit{full}$-$\textit{attn} and \textit{online}$-$\textit{attn} models use mean pooling instead of attention to concatenate features, i.e.~it sets $w_j=1$ in Equation~\ref{eq:context}.  

\noindent \textbf{Captioning results.} Since all the previous work has focused on captioning complete videos, We find that \textit{LSTM-YT} performs much worse than other models as it tries to encode long sequences of video by mean pooling their features (Table~\ref{tab:multicaption}). \textit{H-RNN} performs slightly better but attends over object level features to generate sentence, which causes it to only slightly outperform \textit{LSTM-YT} since we demonstrated earlier that the captions in our dataset are not object centric but action centric instead. \textit{S2VT} and our \textit{no context} model performs better than the previous baselines with a CIDEr score of $20.97$ as it uses an RNN to encode the video features. We see an improvement in performance to $22.19$ and $22.94$ when we incorporate context from past events into our \textit{online}$-$\textit{attn} and \textit{online} models. Finally, we also considering events that will happen in the future, we see further improvements to $24.24$ and $24.56$ for the \textit{full}$-$\textit{attn} and \textit{full} models. Note that while the improvements from using attention is not too large, we see greater improvements amongst videos with more events, suggesting that attention is useful for longer videos.

\begin{figure}[t]
  \centering
   \begin{subfigure}[b]{0.48\columnwidth}
        \centering
        \includegraphics[width=\textwidth]{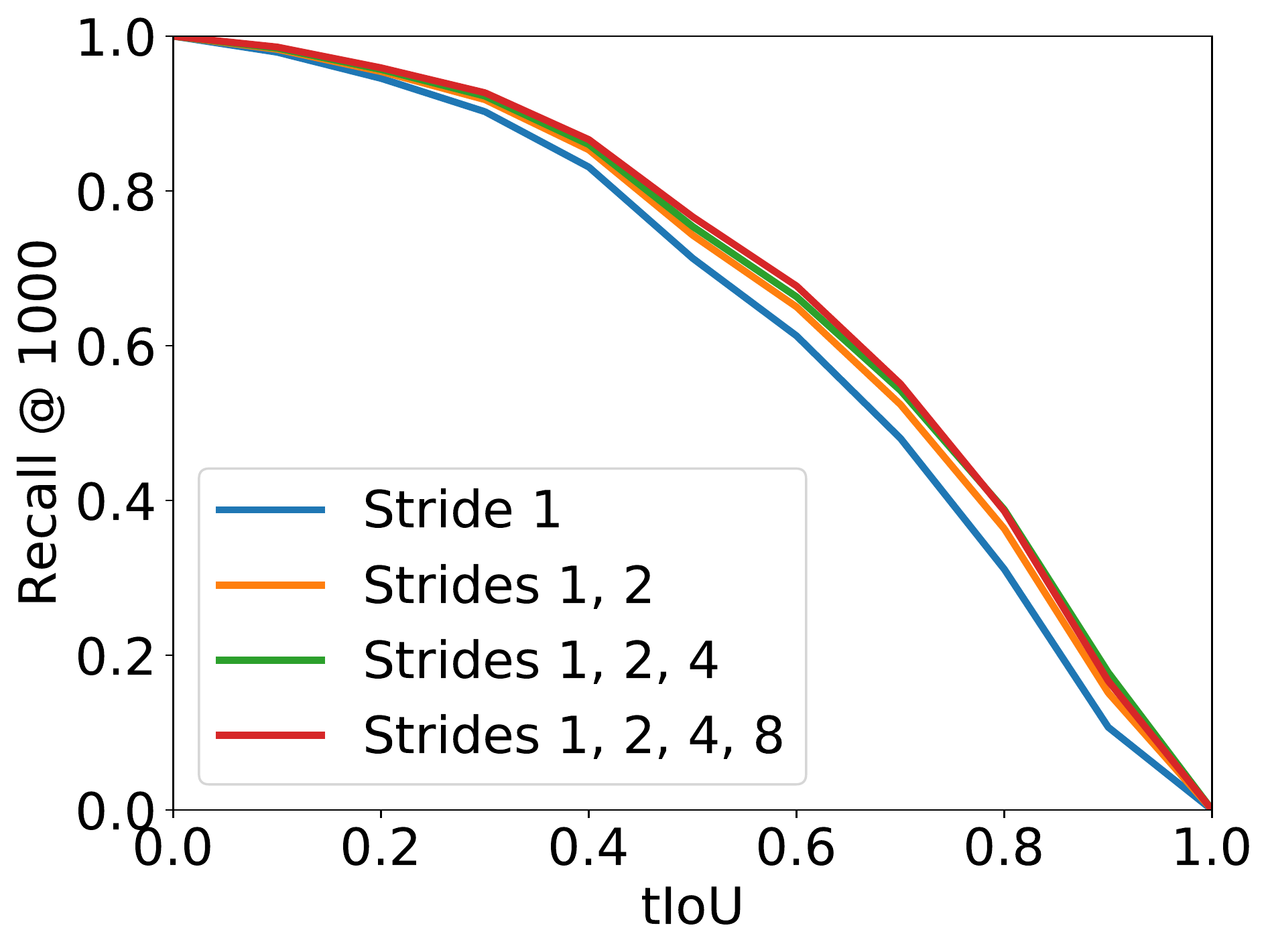}
   \end{subfigure}
   \begin{subfigure}[b]{0.48\columnwidth}
        \centering
        \includegraphics[width=\textwidth]{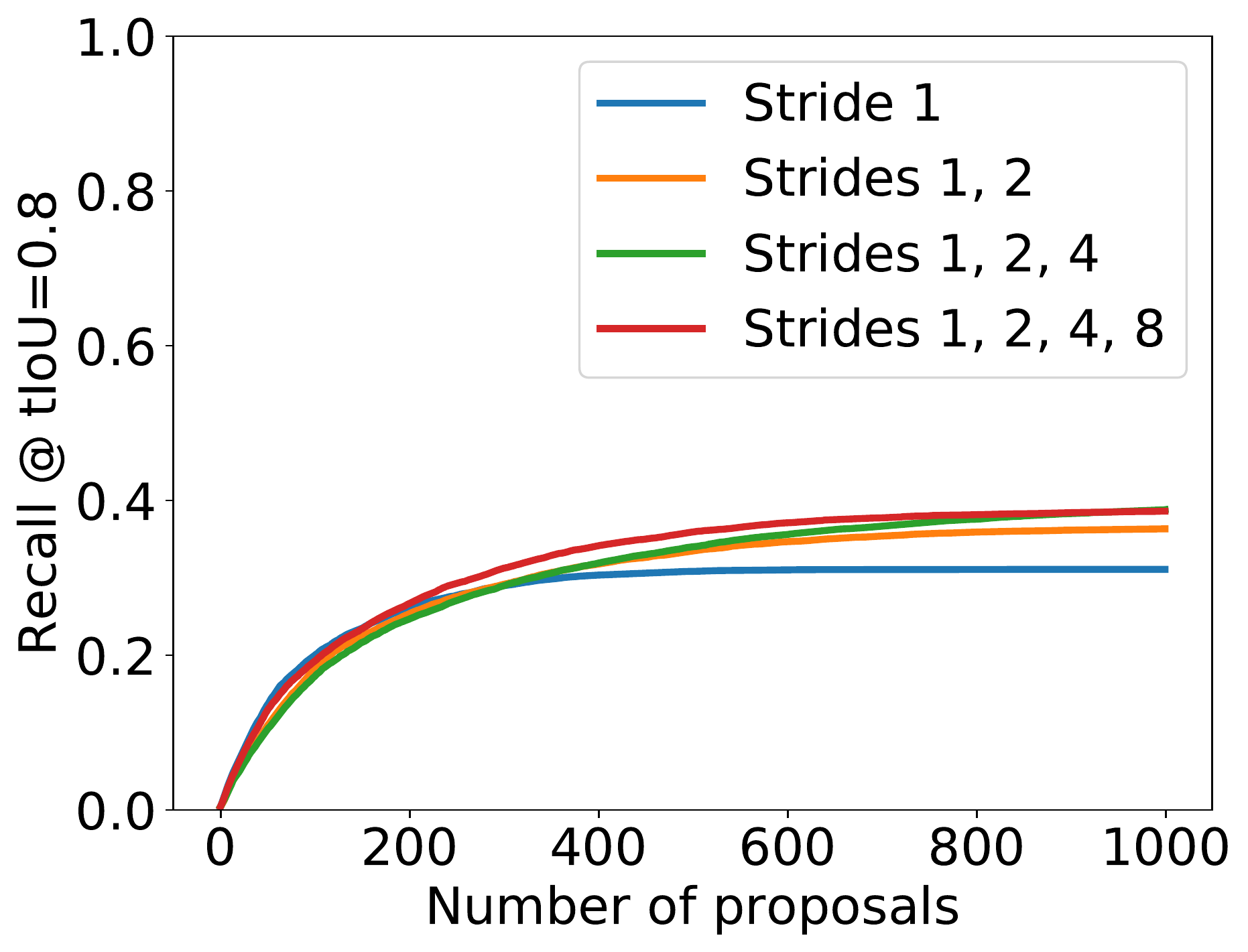}
   \end{subfigure}
  
\caption{Evaluating our proposal module, we find that sampling videos at varying strides does in fact improve the module's ability to localize events, specially longer events.}
\label{fig:iou_localization}
\end{figure}
\noindent \textbf{Sentence order.} To further benchmark the improvements calculated from utilizing past and future context, we report results using ground truth proposals for the first three sentences in each video (Table~\ref{tab:sents}). While there are videos with more than three sentences, we report results only for the first three because almost all the videos in the dataset contains at least three sentences. We notice that the \textit{online} and \textit{full} context models see most of their improvements from subsequent sentences, i.e.~not the first sentence. In fact, we notice that after adding context, the CIDEr score for the \textit{online} and \textit{full} models tend to decrease for the $1^{st}$ sentence.

\begin{table*}[t]
\centering
\begin{tabular}{l c c c c | c c c c}
& \multicolumn{4}{c|}{\textbf{Video retrieval}} & \multicolumn{4}{c}{\textbf{Paragraph retrieval}} \\
                                             & R@1           & R@5           & R@50          & Med. rank & R@1 & R@5 & R@50 & Med. rank\\ \hline \hline
LSTM-YT~\cite{venugopalan2015sequence}       & 0.00          & 0.04          & 0.24          & 102         & 0.00          & 0.07          & 0.38 & 98 \\
no context~\cite{venugopalan2014translating} & 0.05          & 0.14          & 0.32          & 78          & 0.07          & 0.18          & 0.45 & 56 \\
online (ours)                           & 0.10          & \textbf{0.32} & 0.60          & 36          & 0.17          & 0.34          & 0.70 & 33\\
full (ours)                     & \textbf{0.14} & \textbf{0.32} & \textbf{0.65} & \textbf{34} & \textbf{0.18} & \textbf{0.36} & \textbf{0.74} & \textbf{32}
\end{tabular}
  \caption{Results for video and paragraph retrieval. We see that the utilization of context to encode video events help us improve retrieval. R@$k$ measures the recall at varying thresholds $k$ and med. rank measures the median rank the retrieval.}
\label{tab:retrieval}
\end{table*}

\noindent \textbf{Results for dense-captioning events.}  When using proposals instead of ground truth events (Table~\ref{tab:multicaption}), we see a similar trend where adding more context improves captioning. However, we also see that the improvements from attention are more pronounced since there are many events that the model has to caption. Attention allows the model to adequately focus in on select other events that are relevant to the current event. We show examples qualitative results from the variants of our models in Figure~\ref{fig:qualitative_results}. In (a), we see that the last caption in the \textit{no context} model drifts off topic while the \textit{full} model utilizes context to generate more reasonable context. In (c), we see that our \textit{full} context model is able to use the knowledge that the vegetables are later \desc{mixed in the bowl} to also mention \desc{the bowl} in the third and fourth sentences, propagating context back through to past events. However, context is not always successful at generating better captions. In (c), when the proposed segments have a high overlap, our model fails to distinguish between the two events, causing it to repeat captions.

\subsection{Event localization}
\label{sec:localization}
One of the main goals of this paper is to develop models that can locate any given event within a video. Therefore, we test how well our model can predict the temporal location of events within the corresponding video, in isolation of the captioning module. Recall that our variant of the proposal module uses proposes videos at different strides. Specifically, we test with strides of $1$, $2$, $4$ and $8$. Each stride can be computed in parallel, allowing the proposal to run in a single pass.

\noindent \textbf{Setup.} We evaluate our proposal module using recall (like previous work~\cite{escorcia2016daps}) against (1) the number of proposals and (2) the IoU with ground truth events. Specifically, we are testing whether, the use of different strides does in fact improve event localization. 

\noindent \textbf{Results.} Figure~\ref{fig:iou_localization} shows the recall of predicted localizations that overlap with ground truth over a range of IoU's from $0.0$ to $1.0$ and number of proposals ranging till $1000$. We find that using more strides improves recall across all values of IoU's with diminishing returns
. We also observe that when proposing only a few proposals, the model with stride $1$ performs better than any of the multi-stride versions. This occurs because there are more training examples for smaller strides as these models have more video frames to iterate over, allowing them to be more accurate. So, when predicting only a few proposals, the model with stride 1 localizes the most correct events. However, as we increase the number of proposals, we find that the proposal network with only a stride of $1$ plateaus around a recall of $0.3$, while our multi-scale models perform better. 

\subsection{Video and paragraph retrieval}
\label{sec:retrieval}
While we introduce dense-captioning events, a new task to study video understanding, we also evaluate our intuition to use context on a more traditional task: video retrieval.

\noindent \textbf{Setup.} In video retrieval, we are given a set of sentences that describe different parts of a video and are asked to retrieve the correct video from the test set of all videos. Our retrieval model is a slight variant on our dense-captioning model where we encode all the sentences using our captioning module and then combine the context together for each sentence and match each sentence to multiple proposals from a video. We assume that we have ground truth proposals for each video and encode each proposal using the LSTM from our proposal model. We train our model using a max-margin loss that attempts to align the correct sentence encoding to its corresponding video proposal encoding. We also report how this model performs if the task is reversed, where we are given a video as input and are asked to retrieve the correct paragraph from the complete set of paragraphs in the test set.  

\noindent \textbf{Results.} We report our results in Table~\ref{tab:retrieval}. We evaluate retrieval using recall at various thresholds and the median rank. We use the same baseline models as our previous tasks. We find that models that use RNNs (\textit{no context}) to encode the video proposals perform better than max pooling video features (LSTM-YT). We also see a direct increase in performance when context is used. Unlike dense-captioning, we do not see a marked increase in performance when we include context from future events as well. We find that our online models performs almost at par with our full model.

\section{Conclusion}
We introduced the task of dense-captioning events and identified two challenges: (1) events can occur within a second or last up to minutes, and (2) events in a video are related to one another. To tackle both these challenges, we proposed a model that combines a new variant of an existing proposal module with a new captioning module. The proposal module samples video frames at different strides and gathers evidence to propose events at different time scales in one pass of the video. The captioning module attends over the neighboring events, utilizing their context to improve the generation of captions. We compare variants of our model and demonstrate that context does indeed improve captioning. 
We further show how the captioning model uses context to improve video retrieval and how our proposal model uses the different strides to improve event localization. Finally, this paper also releases a new dataset for dense-captioning events: \datasetname.

\fi

\ifsupple
\section{Supplementary material}
In the supplementary material, we compare and contrast our dataset with other datasets and provide additional details about our dataset. We include screenshots of our collection interface with detailed instructions. We also provide additional details about the workers who completed our tasks.

\begin{figure}[t]
  \centering
  \includegraphics[width=\columnwidth]{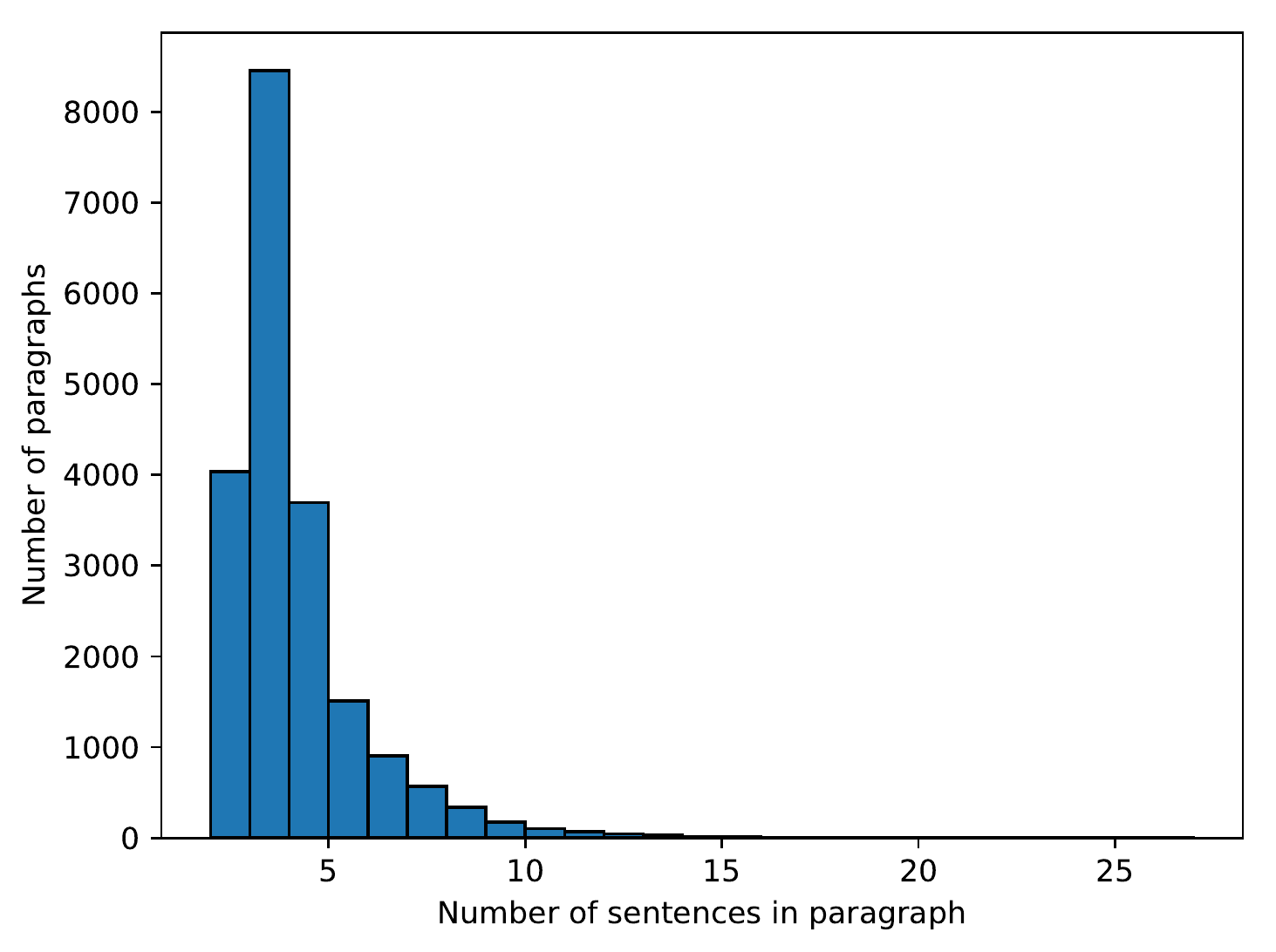}
\caption{The number of sentences within paragraphs is normally distributed, with on average \averageSentencesPerVideo sentences per paragraph.}
\label{fig:sen_para}
\end{figure}  

\subsection{Comparison to other datasets.}
Curation and open distribution is closely correlated with progress in the field of video understanding (Table~\ref{tab:datasets}). The KTH dataset~\cite{schuldt2004recognizing} pioneered the field by studying human actions with a black background. Since then, datasets like UCF101~\cite{soomro2012ucf101}, Sports 1M~\cite{karpathy2014large}, Thumos 15~\cite{THUMOS15} have focused on studying actions in sports related internet videos while HMDB 51~\cite{kuehne2011hmdb} and Hollywood 2~\cite{marszalek09} introduced a dataset of movie clips. Recently, ActivityNet~\cite{caba2015activitynet} and Charades~\cite{sigurdsson2016hollywood} broadened the domain of activities captured by these datasets by including a large set of human activities. In an effort to map video semantics with language, MPII MD~\cite{rohrbach15cvpr} and M-VAD~\cite{torabi2015using} released short movie clips with descriptions. In an effort to capture longer events, MSR-VTT~\cite{xu2016msr}, MSVD~\cite{chen:acl11} and YouCook~\cite{DaXuDoCVPR2013} collected a dataset with slightly longer length, at the cost of a few descriptions than previous datasets. To further improve video annotations, KITTI~\cite{Geiger2013IJRR} and TACoS~\cite{tacos:regnerietal:tacl} also temporally localized their video descriptions. Orthogonally, in an effort to increase the complexity of descriptions, TACos multi-level~\cite{rohrbach2014coherent} expanded the TACoS~\cite{tacos:regnerietal:tacl} dataset to include paragraph descriptions to instructional cooking videos. However, their dataset is constrained in the ``cooking'' domain and contains in the order of a $100$ videos, making it unsuitable for dense-captioning of events as the models easily overfit to the training data.

Our dataset, \datasetname, aims to bridge these three orthogonal approaches by temporally annotating long videos while also building upon the complexity of descriptions. \datasetname~ contains videos that an average of 180s long with the longest video running to over 10 minutes. It contains a total of \totalSentences sentences, where each sentence is temporally localized. Unlike TACoS multi-level, we have two orders of magnitude more videos and provide annotations for an open domain. Finally, we are also the first dataset to enable the study of concurrent events, by allowing our events to overlap. 

\begin{table*}[t]
\centering
\begin{tabular}{l c c l c c c c c}
 Dataset                                        & Domain  & \# videos  & Avg. length & \# sentences    & Des.       & Loc. Des.  & paragraphs & overlapping  \\ \hline \hline
 UCF101~\cite{soomro2012ucf101}                 & sports  & 13k        & 7s          & -               & -          & -          & -          & - \\
 Sports 1M~\cite{karpathy2014large}             & sports  & 1.1M       & 300s        & -               & -          & -          & -          & - \\
 Thumos 15~\cite{THUMOS15}                      & sports  & 21k        & 4s          & -               & -          & -          & -          & - \\
 HMDB 51~\cite{kuehne2011hmdb}                  & movie   & 7k         & 3s          & -               & -          & -          & -          & - \\
 Hollywood 2~\cite{marszalek09}                 & movie   & 4k         & 20s         & -               & -          & -          & -          & - \\
 MPII cooking~\cite{rohrbach2012database}       & cooking & 44         & \textbf{600s}        & -               & -          & -          & -          & - \\
 ActivityNet~\cite{caba2015activitynet}         & human   & 20k        & \textbf{180s}        & -               & -          & -          & -          & - \\
 MPII MD~\cite{rohrbach15cvpr}                  & movie   & 68k        & 4s          & 68,375          & \checkmark & -          & -          & - \\
 M-VAD~\cite{torabi2015using}                   & movie   & 49k        & 6s          & 55,904          & \checkmark & -          & -          & - \\
 MSR-VTT~\cite{xu2016msr}                       & \textbf{open}    & \textbf{10k}        & 20s         & \textbf{200,000}         & \checkmark & -          & -          & - \\
 MSVD~\cite{chen:acl11}                         & \textbf{human}   & \textbf{2k}         & 10s         & \textbf{70,028}          & \checkmark & -          & -          & - \\
 YouCook~\cite{DaXuDoCVPR2013}                  & cooking & 88         & -           & 2,688           & \checkmark & -          & -          & - \\
 Charades~\cite{sigurdsson2016hollywood}        & \textbf{human}   & \textbf{10k}        & 30s         & 16,129          & \checkmark & -          & -          & - \\
 KITTI~\cite{Geiger2013IJRR}                    & driving & 21         & 30s         & 520             & \checkmark & \checkmark & -          & - \\
 TACoS~\cite{tacos:regnerietal:tacl}            & cooking & 127        & \textbf{360s}           & 11,796          & \checkmark & \checkmark & -          & - \\
 TACos multi-level~\cite{rohrbach2014coherent}  & cooking & 127        & \textbf{360s}           & 52,593          & \checkmark & \checkmark & \checkmark & - \\ \hline
 \datasetname~ (ours)                            & \textbf{open}    & \textbf{20k}        & \textbf{180s}        & \textbf{\totalSentences} & \checkmark & \checkmark & \checkmark & \checkmark \\
\end{tabular}
\caption{Compared to other video datasets, \datasetname~ contains long videos with a large number of sentences that are all temporally localized and is the only dataset that contains overlapping events. (Loc. Des. shows which datasets contain temporally localized language descriptions. Bold fonts are used to highlight the nearest comparison of our model with existing models.)}
\label{tab:datasets}
\end{table*}

\subsection{Detailed dataset statistics}
As noted in the main paper, the number of sentences accompanying each video is normally distributed, as seen in Figure~\ref{fig:sen_para}. On average, each video contains $\averageSentencesPerVideo \pm \stdDevSentencesPerVideo$ sentences. Similarly, the number of words in each sentence is normally distributed, as seen in Figure~\ref{fig:words_per_sen}. On average, each sentence contains $\averageWordsPerSentence \pm \stdDevWordsPerSentence$ words, and each video contains $\averageWordsPerParagraph \pm \stdDevWordsPerParagraph$ words.

\begin{figure}[t]
  \centering
  \includegraphics[width=\columnwidth]{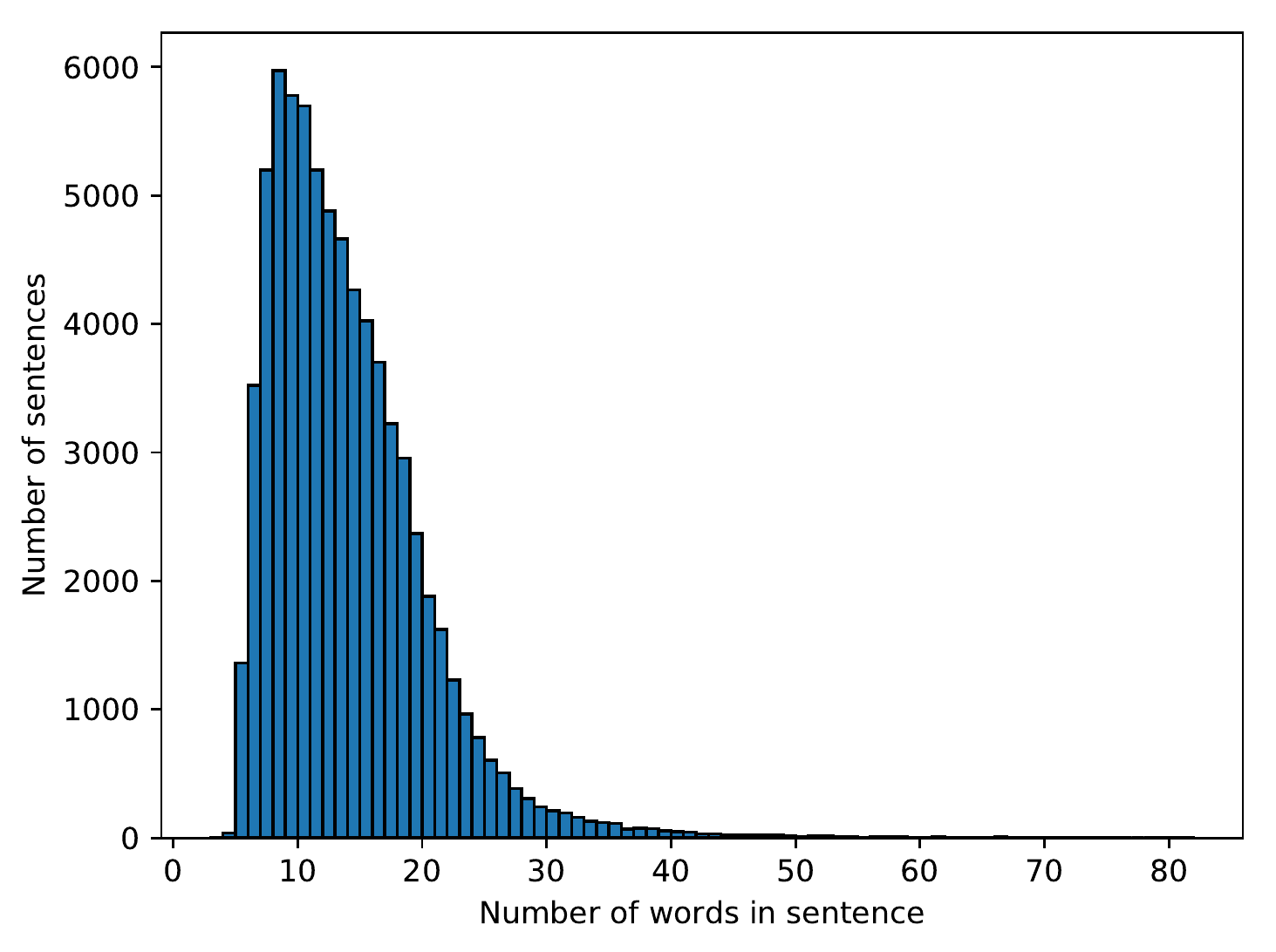}
\caption{The number of words per sentence within paragraphs is normally distributed, with on average \averageWordsPerSentence words per sentence.}
\label{fig:words_per_sen}
\end{figure}

There exists interaction between the video content and the corresponding temporal annotations. In Figure~\ref{fig:sen_vlength}, the number of sentences accompanying a video is shown to be positively correlated with the video's length: each additional minute adds approximately $1$ additional sentence description. Furthermore, as seen in Figure~\ref{fig:time_heatmap}, the sentence descriptions focus on the middle parts of the video more than the beginning or end. 

\begin{figure}[t]
  \centering
  \includegraphics[width=\columnwidth]{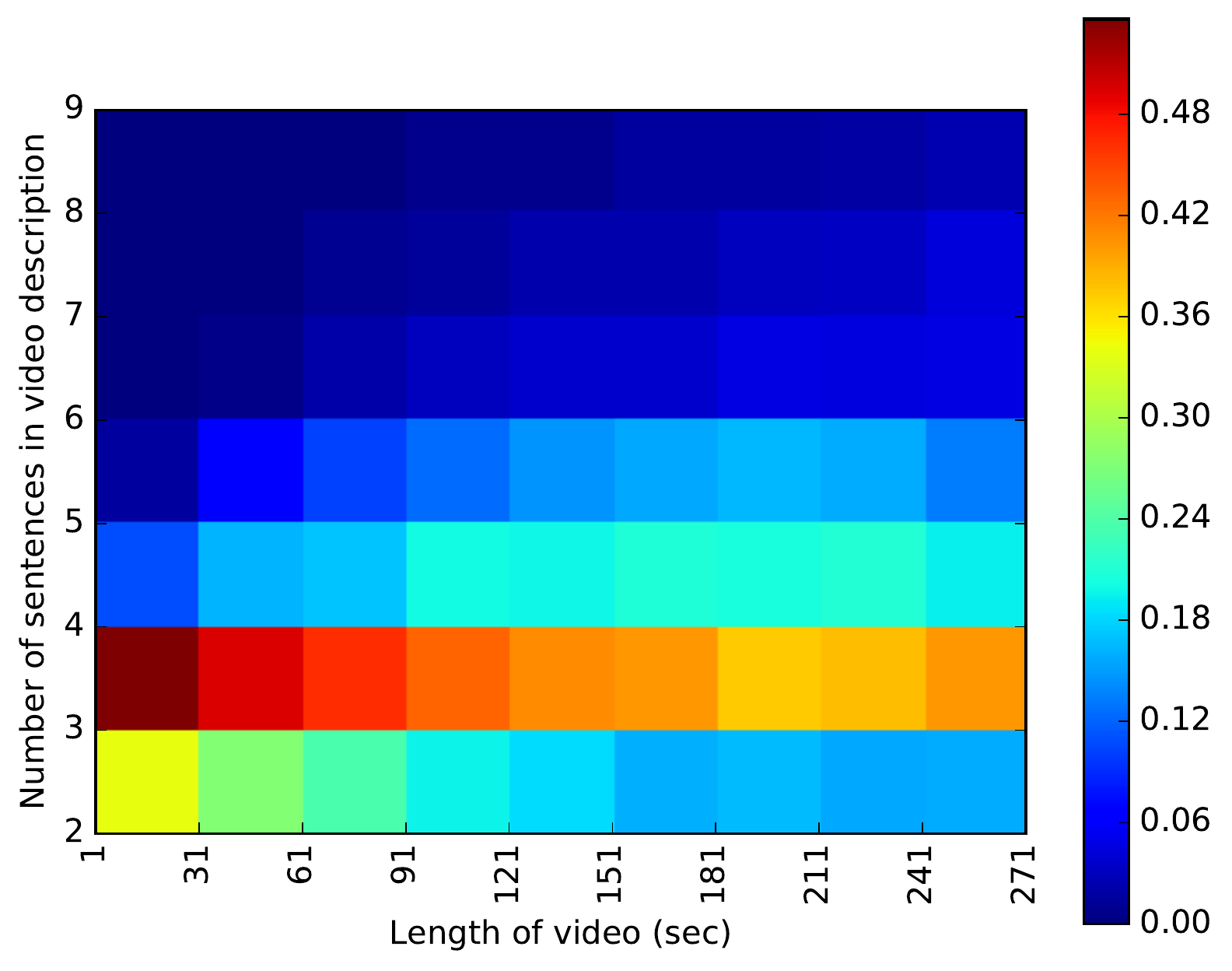}
\caption{Distribution of number of sentences with respect to video length. In general the longer the video the more sentences there are, so far on average each additional minute adds one more sentence to the paragraph.}
\label{fig:sen_vlength}
\end{figure}

\begin{figure}[t]
  \centering
  \includegraphics[width=\columnwidth]{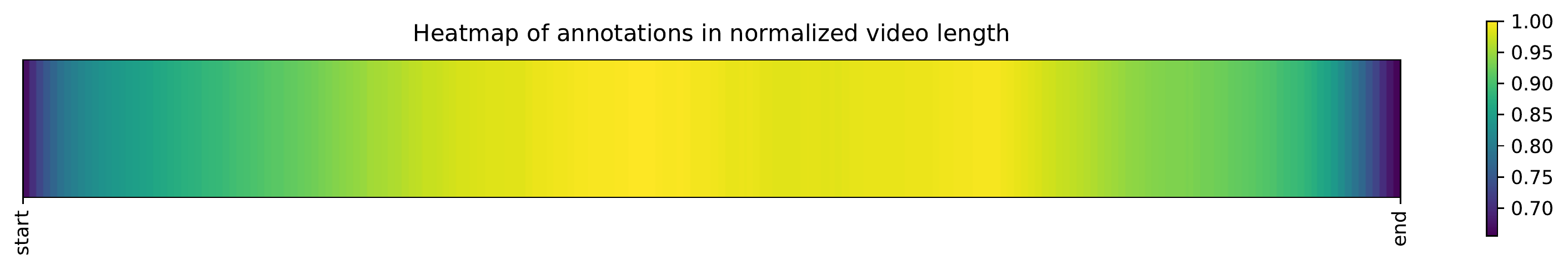}
\caption{Distribution of annotations in time in \datasetname~ videos, most of the annotated time intervals are closer to the middle of the videos than to the start and end.}
\label{fig:time_heatmap}
\end{figure}

When studying the distribution of words in Figures~\ref{fig:top_unigram} and~\ref{fig:top_bigram}, we found that \datasetname~ generally focuses on people and the actions these people take. However, we wanted to know whether \datasetname~ captured the general semantics of the video. To do so, we compare our sentence descriptions against the shorter labels of ActivityNet, since \datasetname~ annotates ActivityNet videos. Figure~\ref{fig:labelpercentage} illustrates that the majority of videos in \datasetname~ often contain ActivityNet's labels in at least one of their sentence descriptions. We find that the many entry-level categories such as \emph{brushing hair} or \emph{playing violin} are extremely well represented by our captions. However, as the categories become more nuanced, such as \emph{powerbocking} or \emph{cumbia}, they are not as commonly found in our descriptions.

\begin{figure}[t]
  \centering
  \includegraphics[width=\columnwidth]{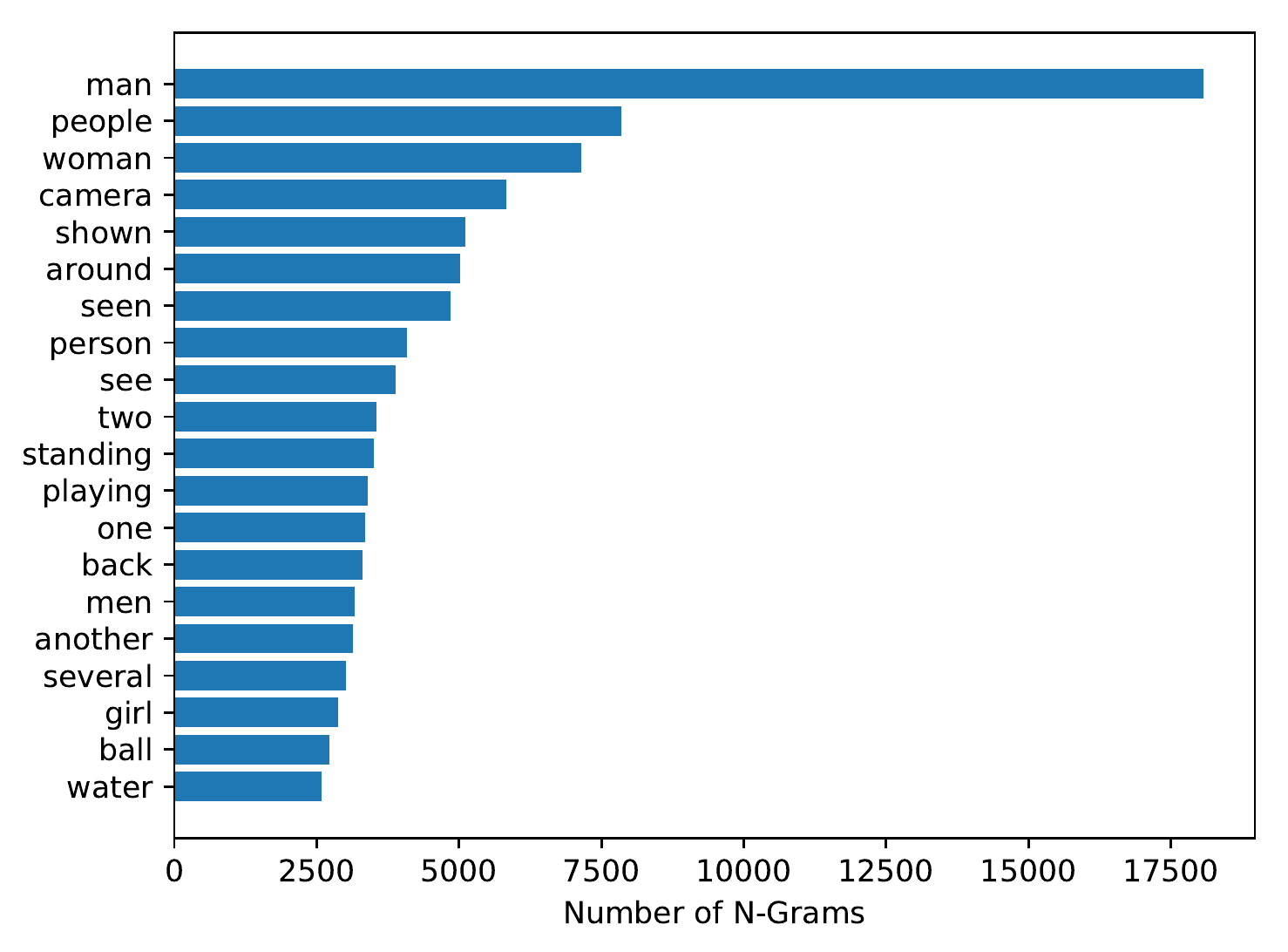}
\caption{The most frequently used words in \datasetname~ with stop words removed.}
\label{fig:top_unigram}
\end{figure}

\begin{figure}[t]
  \centering
 \includegraphics[width=\columnwidth]{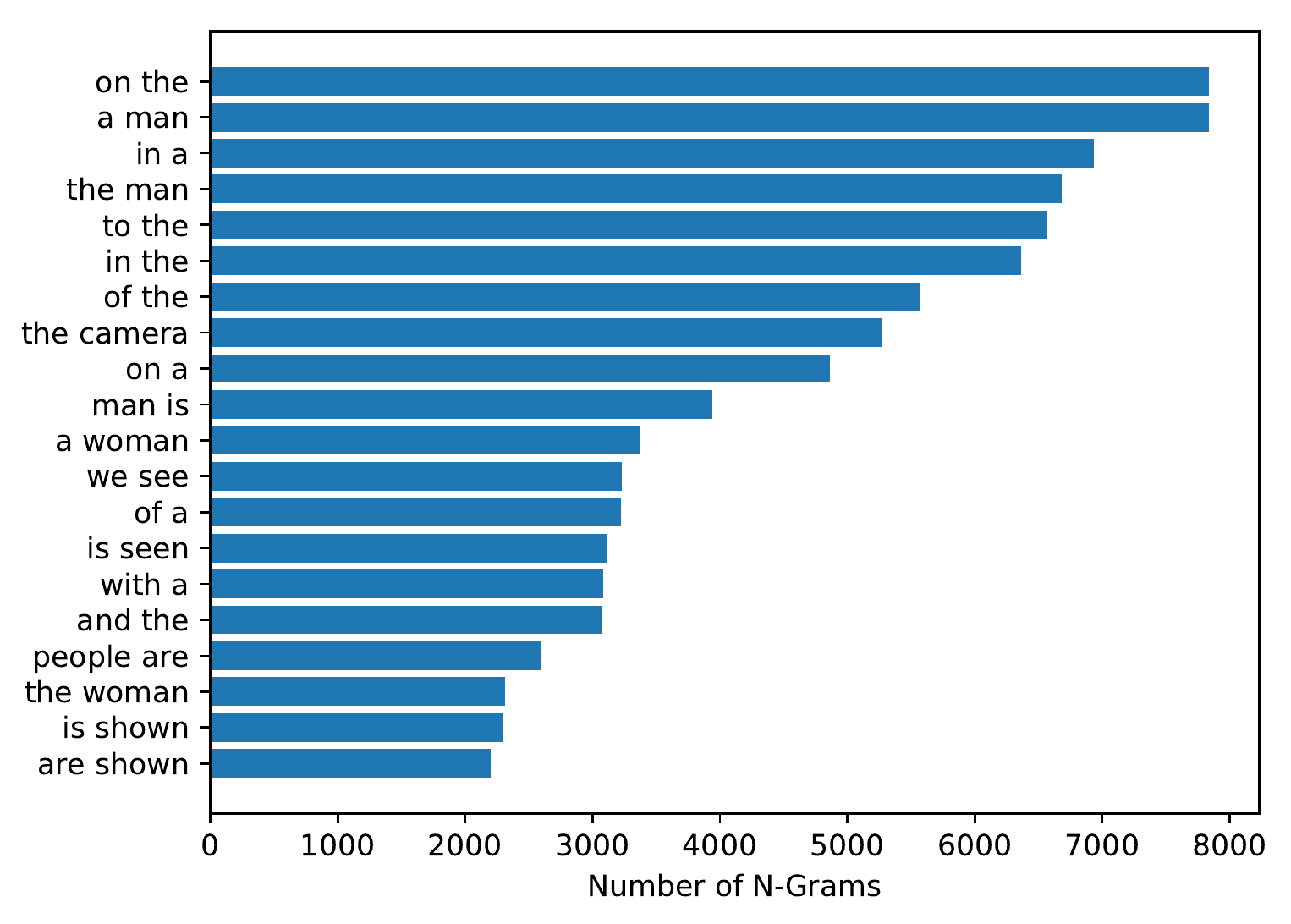}
\caption{The most frequently used bigrams in \datasetname~.}
\label{fig:top_bigram}
\end{figure}

\begin{figure}[t]
  \centering
  \includegraphics[width=\columnwidth]{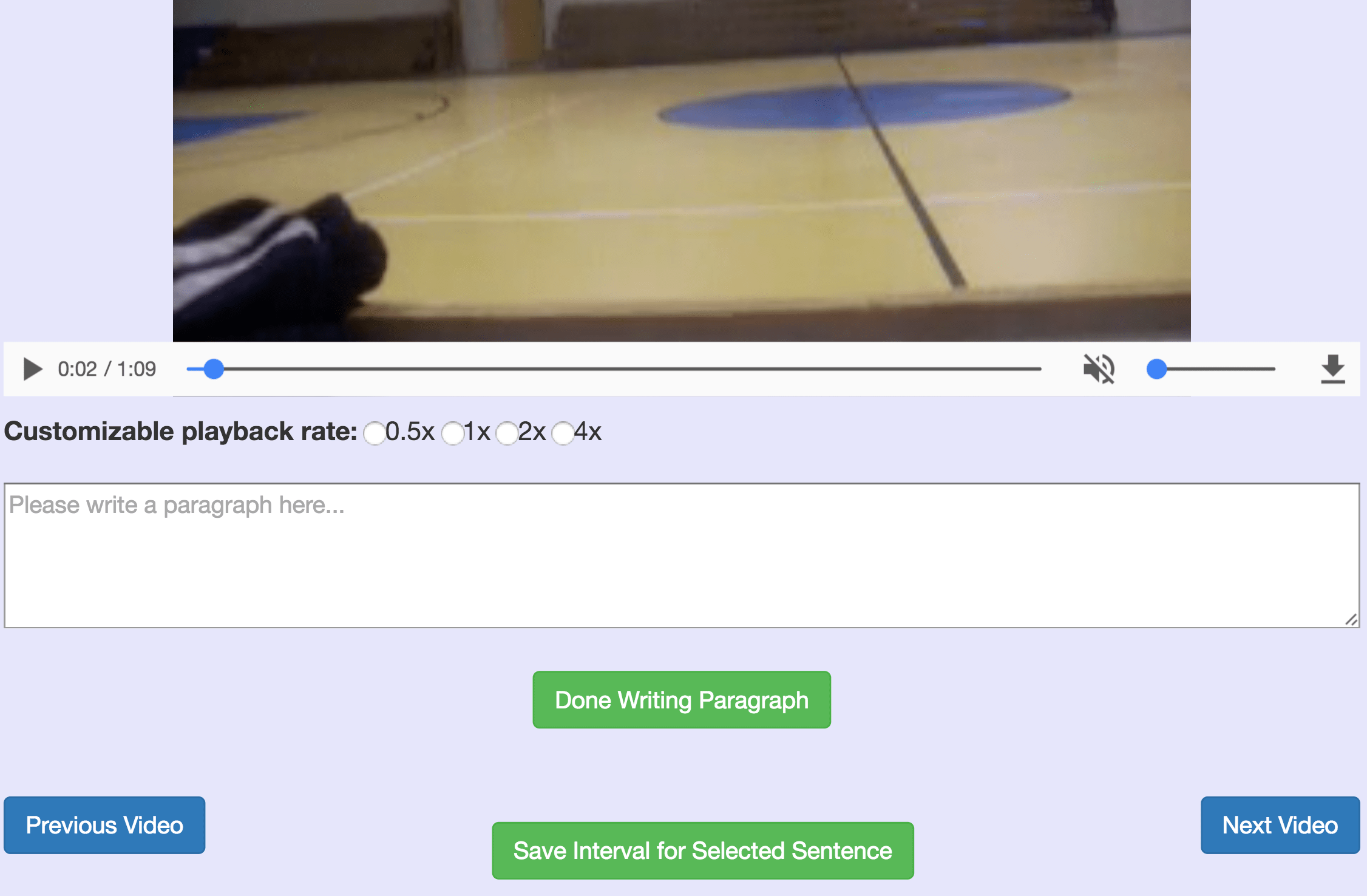}
\caption{Interface when a worker is writing a paragraph. Workers are asked to write a paragraph in the text box and press "Done Writing Paragraph" before they can proceed with grounding each of the sentences.}
\label{fig:interface_paragraph}
\end{figure}

\begin{figure}[t]
  \centering
  \includegraphics[width=\columnwidth]{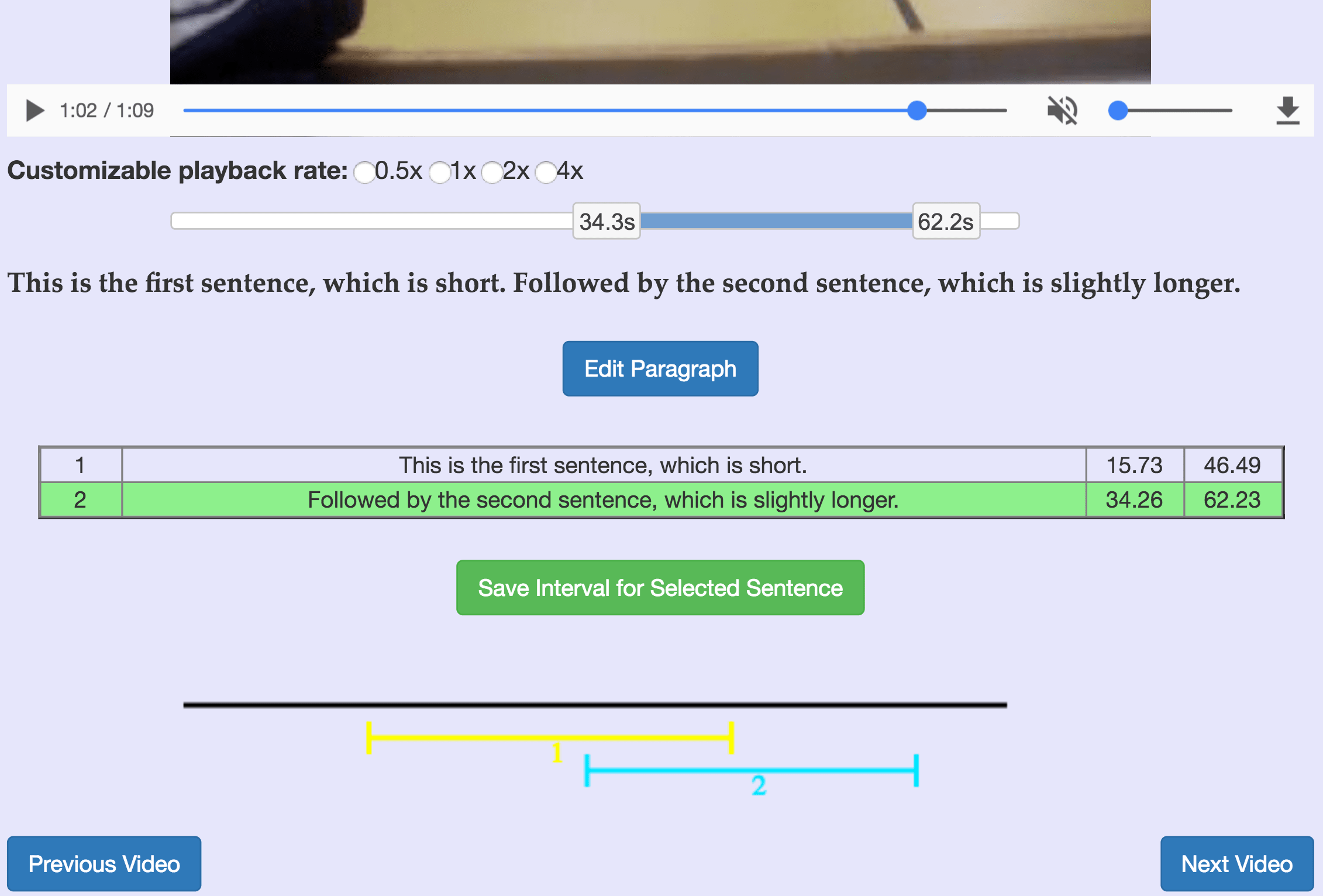}
\caption{Interface when labeling sentences with start and end timestamps. Workers select each sentence, adjust the range slider indicating which segment of the video that particular sentence is referring to. They then click save and proceed to the next sentence.}
\label{fig:interface_timestamps}
\end{figure}

\begin{figure}[t]
  \centering
  \includegraphics[width=\columnwidth]{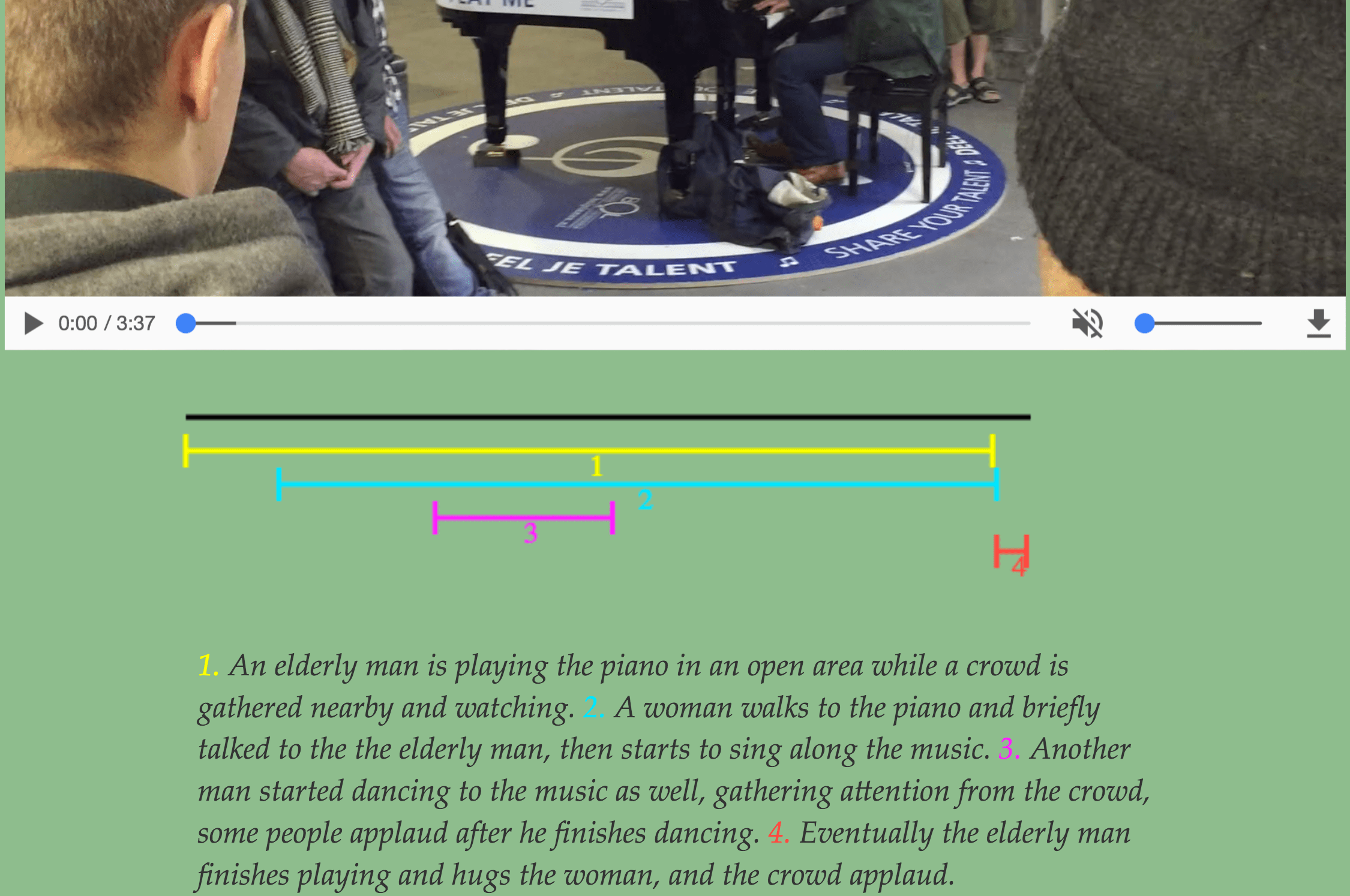}
\caption{We show examples of good and bad annotations to workers. Each task contains one good and one bad example video with annotations. We also explain why the examples are considered to be good or bad.}
\label{fig:interface_intro}
\end{figure}

\subsection{Dataset collection process}
We used Amazon Mechanical Turk to annotate all our videos. Each annotation task was divided into two steps: (1) Writing a paragraph describing all major events happening in the videos in a paragraph, with each sentence of the paragraph describing one event (Figure~\ref{fig:interface_paragraph}; and (2) Labeling the start and end time in the video in which each sentence in the paragraph event occurred (Figure~\ref{fig:interface_timestamps}. We find complementary evidence that workers are more consistent with their video segments and paragraph descriptions if they are asked to annotate visual media (in this case, videos) using natural language first~\cite{krishnavisualgenome}. Therefore, instead of asking workers to segment the video first and then write individual sentences, we asked them to write paragraph descriptions first.

Workers are instructed to ensure that their paragraphs are at least $3$ sentences long where each sentence describes events in the video but also makes a grammatically and semantically coherent paragraph. They were allowed to use co-referencing words (ex, \desc{he}, \desc{she}, etc.) to refer to subjects introduced in previous sentences. We also asked workers to write sentences that were at least $5$ words long. We found that our workers were diligent and wrote an average of \averageWordsPerSentence number of words per sentence. Each of the task and examples (Figure~\ref{fig:interface_intro}) of good and bad annotations.

Workers were presented with examples of good and bad annotations with explanations for what constituted a good paragraph, ensuring that workers saw concrete evidence of what kind of work was expected of them (Figure~\ref{fig:interface_intro}). We paid workers $\$3$ for every $5$ videos that were annotated. This amounted to an average pay rate of $\$8$ per hour, which is in tune with fair crowd worker wage rate~\cite{salehi2015we}.

\subsection{Annotation details}

Following research from previous work that show that crowd workers are able to perform at the same quality of work when allowed to video media at a faster rate~\cite{krishna2016embracing}, we show all videos to workers at $2$X the speed, i.e.~the videos are shown at twice the frame rate. Workers do, however, have the option to watching the videos at the original video speed and even speed it up to $3$X or $4$X the speed. We found, however, that the average viewing rate chosen by workers was $1.91$X while the median rate was $1$X, indicating that a majority of workers preferred watching the video at its original speed. We also find that workers tend to take an average of $2.88$ and a median of $1.46$ times the length of the video in seconds to annotate.

At any given time, workers have the ability to edit their paragraph, go back to previous videos to make changes to their annotations. They are only allowed to proceed to the next video if this current video has been completely annotated with a paragraph with all its sentences timestamped. Changes made to the paragraphs and timestamps are saved when "previous video or "next video" are pressed, and reflected on the page. Only when all videos are annotated can the worker submit the task. In total, we had $112$ workers who annotated all our videos.

\fi

\noindent \textbf{Acknowledgements.}  This research was sponsored in part by grants from the Office of Naval Research (N00014-15-1-2813) and Panasonic, Inc. We thank JunYoung Gwak, Timnit Gebru, Alvaro Soto, and Alexandre Alahi for their helpful comments and discussion.

\small
\bibliographystyle{ieee}
\bibliography{egbib}

\ifsupple
\begin{figure*}[t]
  \centering
  \begin{subfigure}[a]{\textwidth}
        \centering
        \includegraphics[width=.8\textwidth]{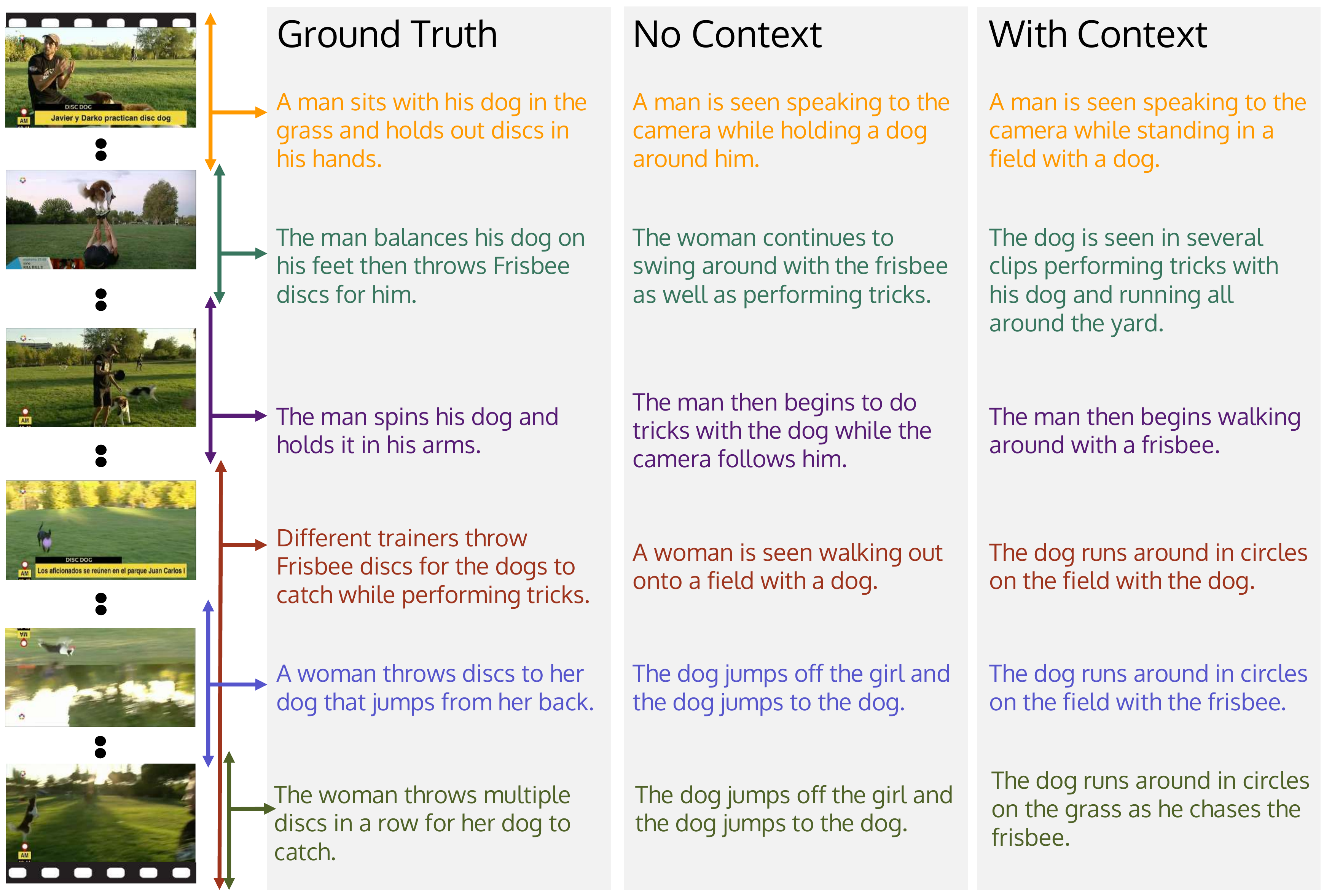}
   \end{subfigure}
   \begin{subfigure}[b]{\textwidth}
        \centering
        \includegraphics[width=.8\textwidth]{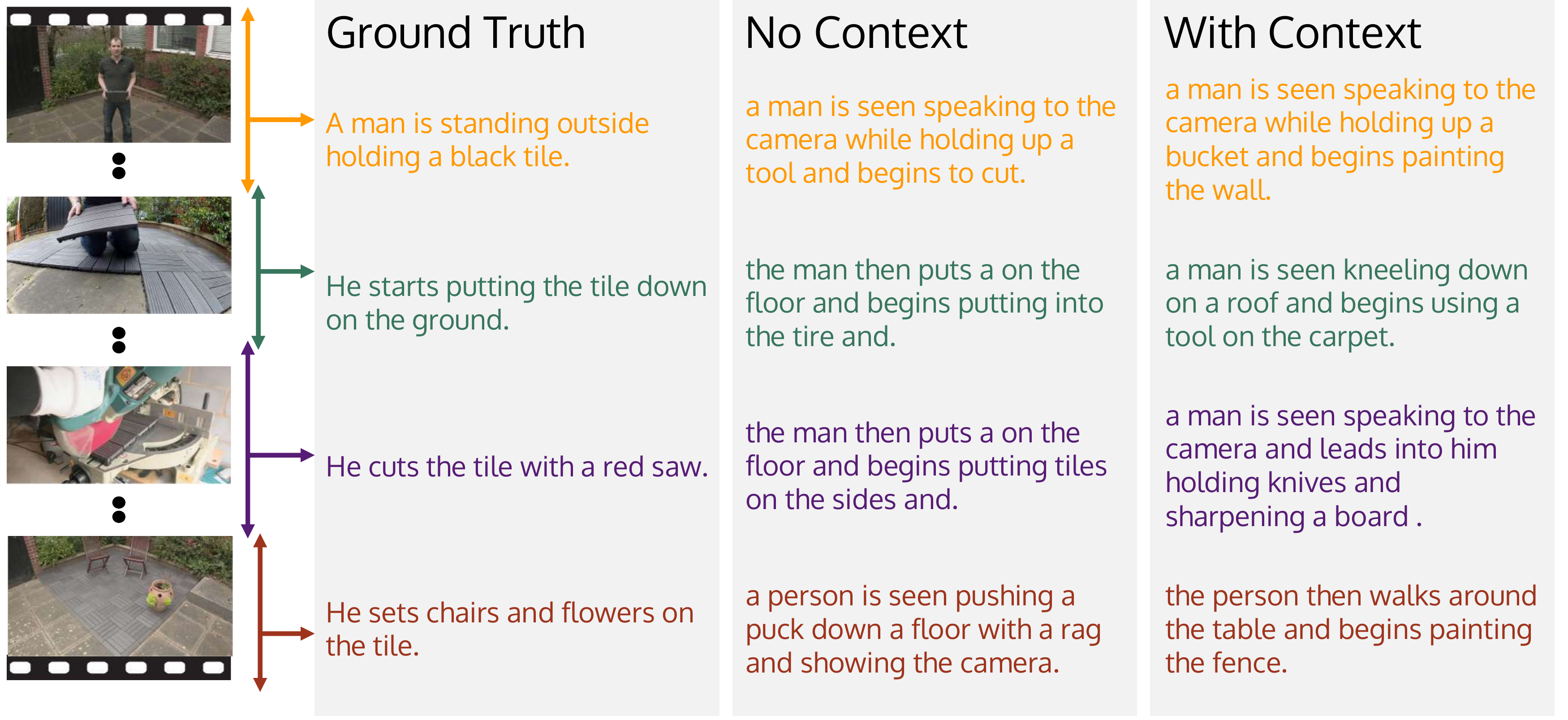}
   \end{subfigure}
   \begin{subfigure}[c]{\textwidth}
        \centering
        \includegraphics[width=.8\textwidth]{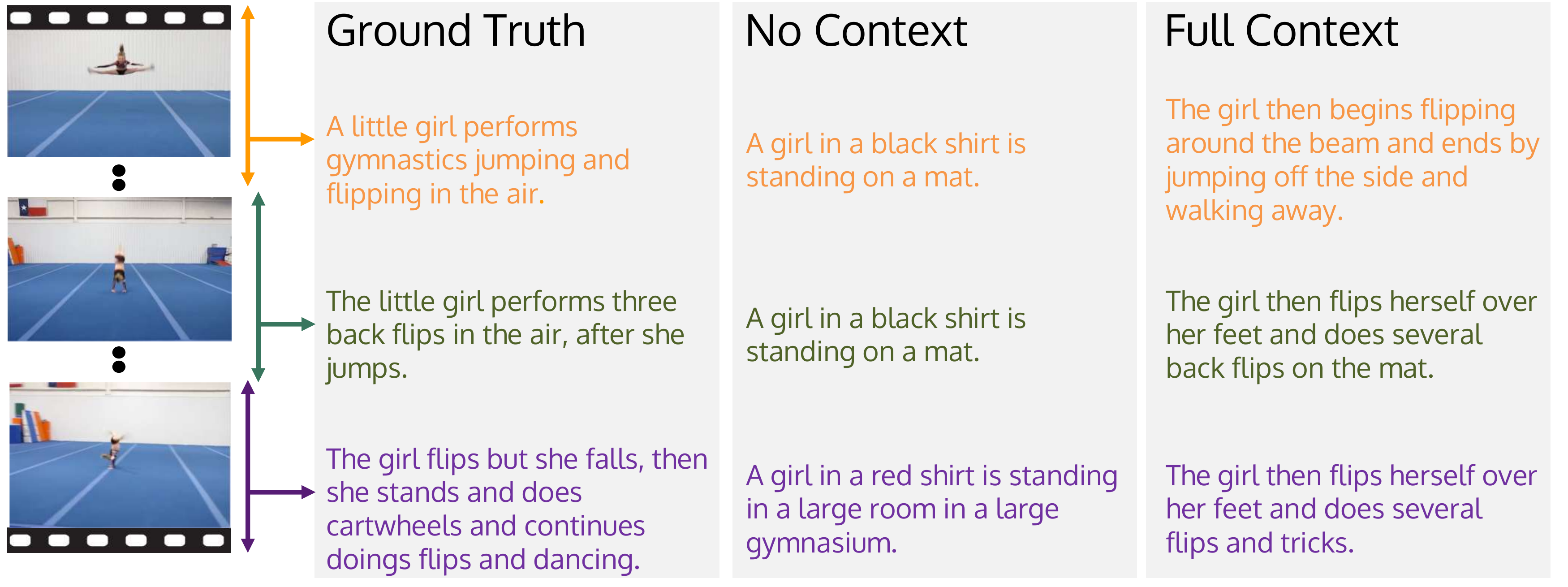}
   \end{subfigure}
  
\caption{More qualitative dense-captioning captions generated using our model. We show captions with the highest overlap with ground truth captions.}
\label{fig:qualitative_results2}
\end{figure*}

\begin{figure*}[t]
  \centering
  \includegraphics[width=0.90\textwidth]{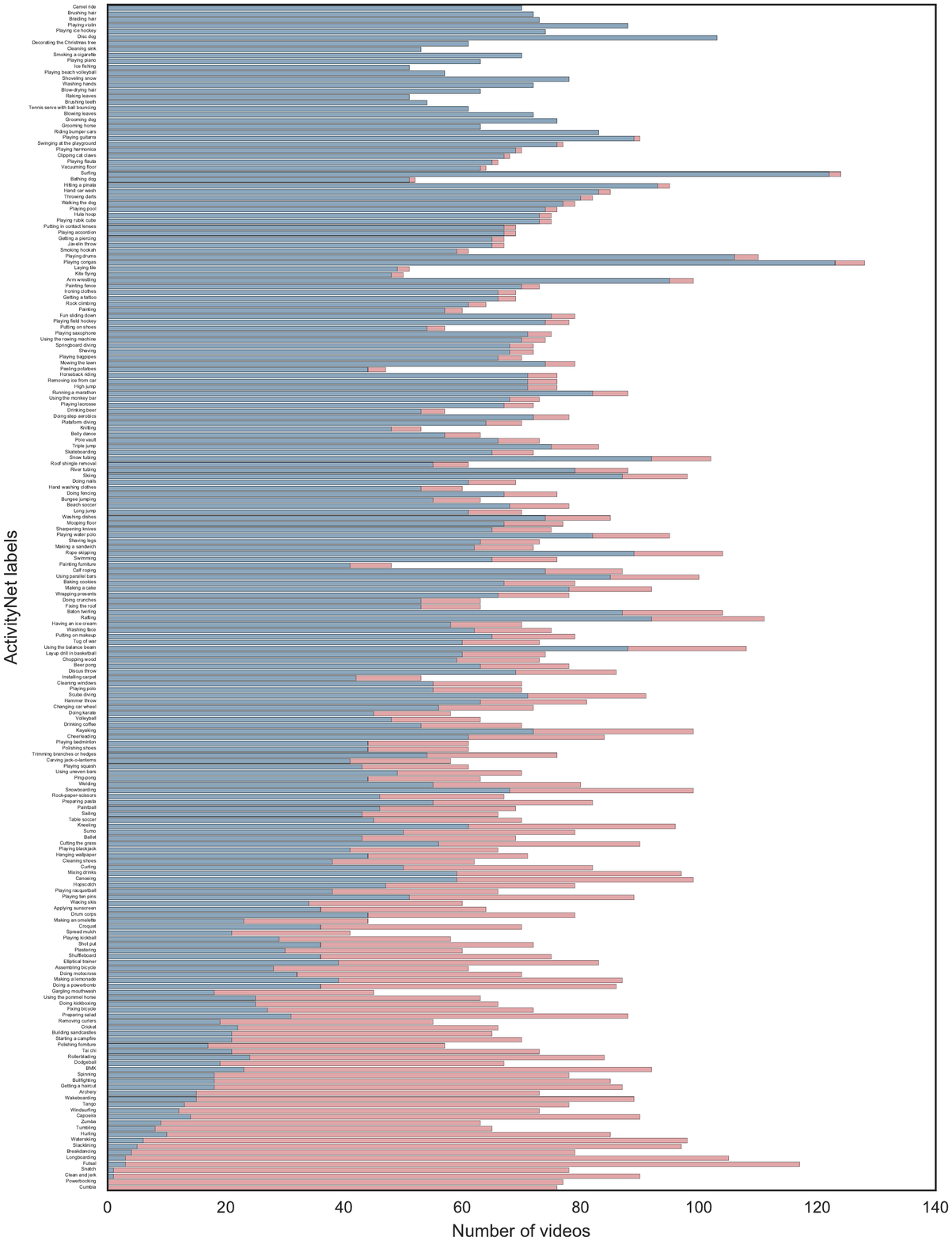}
\caption{The number of videos (red) corresponding to each ActivityNet class label, as well as the number of videos (blue) that has the label appearing in their \datasetname~ paragraph descriptions.}
\label{fig:labelpercentage}
\end{figure*}
\fi
\end{document}